%% file: main.tex
\documentclass[10pt,twocolumn,letterpaper]{article}
\usepackage[misc]{ifsym} 
\usepackage{floatrow}
\usepackage[pagenumbers]{cvpr} 
\usepackage{graphicx}
\usepackage{amsmath}
\usepackage{amssymb}
\usepackage{booktabs}
\usepackage{comment}
\usepackage{xcolor}
\usepackage{multirow, multicol}
\usepackage{floatrow}
\newfloatcommand{capbtabbox}{table}[][\FBwidth]
\usepackage{blindtext}
\usepackage[normalem]{ulem}

\definecolor{citecolor}{RGB}{65,105,225}
\usepackage[pagebackref=true,breaklinks=true,letterpaper=true,colorlinks,
citecolor=citecolor,bookmarks=false]{hyperref}

\usepackage[capitalize]{cleveref}
\crefname{section}{Sec.}{Secs.}
\Crefname{section}{Section}{Sections}
\Crefname{table}{Table}{Tables}
\crefname{table}{Tab.}{Tabs.}

\newcommand{\specialcell}[2][c]{%
  \begin{tabular}[#1]{@{}c@{}}#2\end{tabular}}
\usepackage{pifont} 
\newcommand{\cmark}{\ding{51}}
\newcommand{\xmark}{\ding{55}}

\begin{document}

\title{StyleGAN-Human: A Data-Centric Odyssey of Human Generation}

\author{
    Jianglin Fu$^{1*}$
    \quad
    Shikai Li$^{1*}$
    \quad
    Yuming Jiang$^{2}$
    \quad
    Kwan-Yee Lin$^{1}$ 
    \quad
    Chen Qian$^{1}$ \\
    \quad
    Chen Change Loy$^{2}$
    \quad
    Wayne Wu\textsuperscript{1,3 \Letter} 
    \quad
    Ziwei Liu$^{2}$ \\
    $^{1}$ SenseTime Research
    \quad
    $^{2}$ S-Lab, Nanyang Technological University
    \quad
    $^{3}$ Shanghai AI Laboratory
    \\
    {\tt\small \{fujianglin,lishikai,linjunyi,qianchen\}@sensetime.com} 
    \\
    {\tt\small \{yuming002, ccloy, ziwei.liu\}@ntu.edu.sg}
    \\
    {\tt\small wuwenyan0503@gmail.com}
}

\twocolumn[{
            \renewcommand\twocolumn[1][]{#1}
            \vspace{-1em}
            \maketitle
            \vspace{-3em}
            \begin{center}
                \centering
                \includegraphics[width=0.99\textwidth]{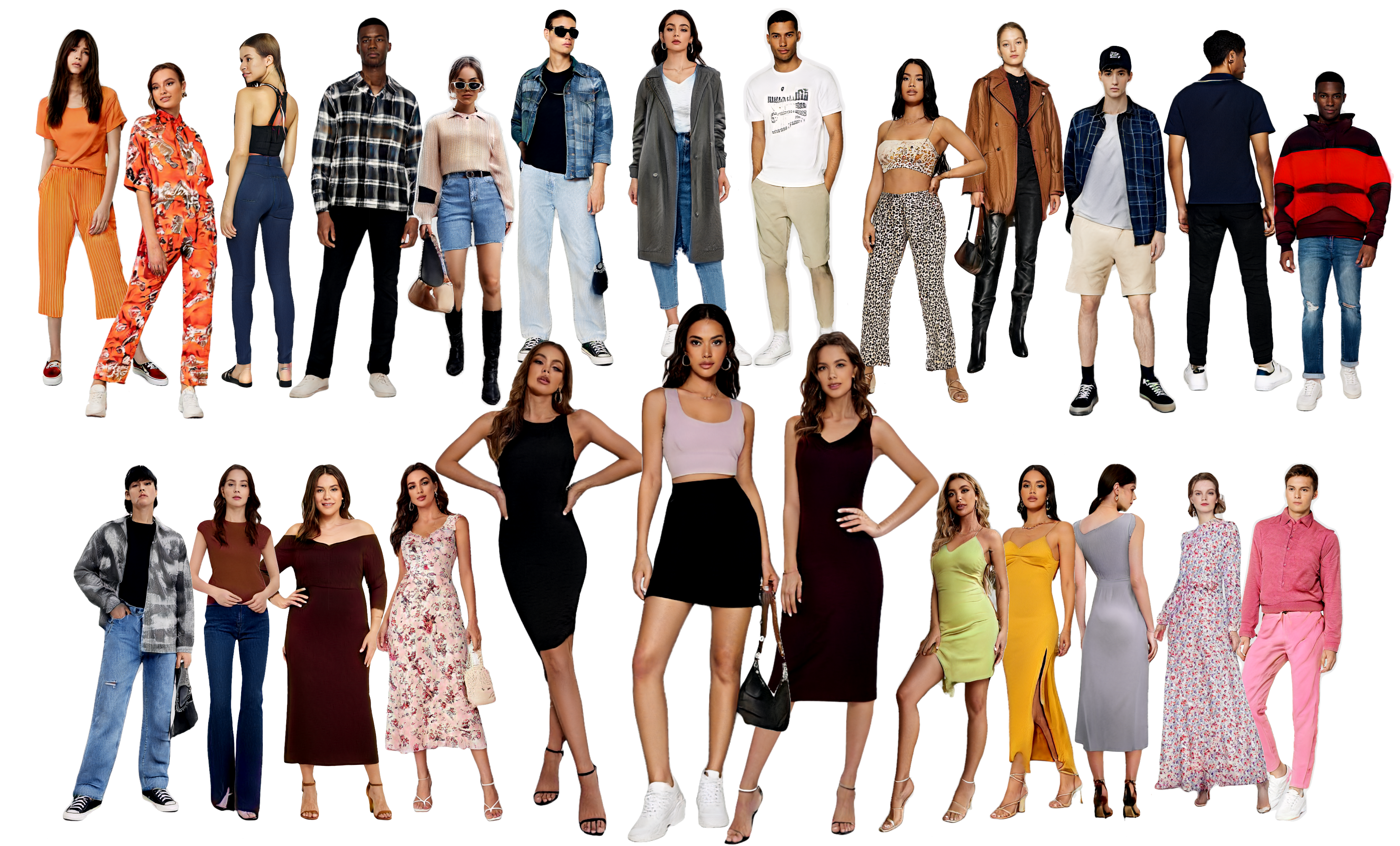}
                \captionof{figure}{\textbf{A Data-Centric Odyssey of Human Generation.} With good ``data engineering'' practices, given a random latent code $z$, the StyleGAN-Human model could generate high-resolution photo-realistic human images as presented. Zoom in for the best view.}
                \label{fig:generatedsample}
            \end{center}
        }]

\def\thefootnote{*}\footnotetext{Equal contribution.}\def\thefootnote{\arabic{footnote}}

\begin{abstract}
Unconditional human image generation is an important task in vision and graphics, enabling various applications in the creative industry. Existing studies in this field mainly focus on ``network engineering'' such as designing new components and objective functions. This work takes a data-centric perspective and investigates multiple critical aspects in ``data engineering'', which we believe would complement the current practice. To facilitate a comprehensive study, we collect and annotate a large-scale human image dataset with over $230K$ samples capturing diverse poses and textures. 
Equipped with this large dataset, we rigorously investigate three essential factors in data engineering for StyleGAN-based human generation, namely data size, data distribution, and data alignment.
Extensive experiments reveal several valuable observations \textit{w.r.t.} these aspects: 
1) Large-scale data, more than $40K$ images, are needed to train a high-fidelity unconditional human generation model with a vanilla StyleGAN.
2) A balanced training set helps improve the generation quality with rare face poses compared to the long-tailed counterpart, whereas simply balancing the clothing texture distribution does not effectively bring an improvement.
3) Human GAN models that employ body centers for alignment outperform models trained using face centers or pelvis points as alignment anchors. 
In addition, a model zoo and human editing applications are demonstrated to facilitate future research in the community. Code and models are publicly available\footnote{Project page: \url{https://stylegan-human.github.io/} \\
Code and models: \url{https://github.com/stylegan-human/StyleGAN-Human}}.

\end{abstract}


\input{section/introduction}
\input{section/related_work}
\input{section/dataset}

\input{section/investigation}
\input{section/model_zoo}

\input{section/future_work}

\input{section/conclusion}


{\small
\bibliographystyle{ieee_fullname}
\bibliography{references}
}

\clearpage

\input{section/appendix}

\end{document}

%% file: section/introduction.tex
\section{Introduction}
\label{sec:intro}
Generating photo-realistic images of clothed humans unconditionally can provide great support for downstream tasks such as human motion transfer~\cite{chan2019everybody,liquidwarpinggan}, digital human animation~\cite{li2021animated}, fashion recommendation~\cite{kang2017visually,Lei_2016_CVPR}, and virtual try-on~\cite{Multi-Pose_tryon,unpair_tryon,Wang_2018_ECCV}. Traditional methods create dressed humans with classical graphics modeling and rendering processes~\cite{gahan20123ds,jojic1999computer,ma2020learning,patel2020tailornet,pumarola20193dpeople,sarkar2020neural,song20163d,xu2020ghum}. Although impressive results have been achieved, these prior works are easy to suffer from the limitation of robustness and generalizability in complex environments. Recent years, Generative Adversarial Networks (GANs) have demonstrated remarkable abilities in real-world scenarios, generating diverse and realistic images by learning from large-quantity and high-quality datasets.~\cite{wgan-gp,pggan,stylegan,dcgan}.

Among the GAN family, StyleGAN2~\cite{stylegan2} stands out in generating faces and simple objects with unprecedented image quality. A major driver behind recent advancements~\cite{abdal2021styleflow,stylegan2ada,stylegan2,StyleRig,wu2021stylespace} on such StyleGAN architectures is the prosperous discovery of ``network engineering'' like designing new components~\cite{stylegan2ada,abdal2021styleflow,wu2021stylespace} and loss functions~\cite{stylegan2,StyleRig}. While these approaches show compelling results in generating diverse objects (\eg, faces of humans and animals), applying them to the photo-realistic generation of articulated humans in natural clothing is still a challenging and open problem. 

In this work, we focus on the task of Unconditional Human Generation, with a specific aim to train a good StyleGAN-based model for articulated humans from a \textit{data-centric} perspective. First, to support the data-centric investigation, collecting a large-scale, high-quality, and diverse dataset of human bodies in clothing is necessary. We propose the Stylish-Humans-HQ Dataset (SHHQ), which contains $230K$ clean full-body images with a resolution of $1024\times512$ at least and up to $2240\times1920$. The SHHQ dataset lays the foundation for extensive experiments on unconditional human generation. Second, based on the proposed SHHQ dataset, we investigate three fundamental and critical questions that were not thoroughly discussed in prior works and attempt to provide useful insights for future research on unconditional human generation.

To extract the questions that are indeed \textit{important} for the community of Unconditional Human Generation, we make an extensive survey on recent literature in the field of general unconditional generation~\cite{goodfellow2014generative,wgan,wgan-gp,lsgan,pggan,biggan,stylegan}. Based on the survey, three questions that are investigated actively can be concluded as below.
\textbf{Question-1}: What is the relationship between the \textit{data size} and the generation quality? Several previous works~\cite{biggan,stylegan2ada,toutouh2020data,zhao2020differentiable,jiang2021DeceiveD} pointed out that the quantity of training data is the primary factor to determine the strategy for improving image quality in face and other object generation tasks. In this study, we want to examine the minimum quantity of training data required to generate human images of high quality without any extensive ``network engineering'' effort. 
\textbf{Question-2}: What is the relationship between the \textit{data distribution} and the generation quality?
This question has received extensive attention~\cite{tailgan,biascorrection,bagan,fairnessgan,fairgan} and leads to a research topic dealing with data imbalance~\cite{liu2019large}. In this study, we aim to exploit data imbalance problem in the human generation task.
\textbf{Question-3}: What is the relationship between the scheme of \textit{data alignment} and the generation quality? Different alignment schemes applied to uncurated faces~\cite{stylegan,kazemi2014one} and non-rigid objects~\cite{biggan,diffusiongan,styleganxl} show success in enhancing training performance. In this study, we seek a better data alignment strategy for human generation.

Based on the proposed SHHQ dataset and observations from our experiments, we establish a Model Zoo with three widely-adopted unconditional generation models, \textit{i.e.}, StyleGAN~\cite{stylegan}, StyleGAN2~\cite{stylegan2}, and alias-free StyleGAN~\cite{stylegan3}, in both resolution of $1024\times512$ and $512\times256$. Although hundreds of StyleGAN-based studies exist for \textit{face} generation/editing tasks, a high-quality and public model zoo for \textit{human} generation/editing with StyleGAN family is still missing.
The provided model zoo is positioned to complement existing facial model zoo. We believe it has great potentials in many human-centric tasks, \eg, human editing, neural rendering, and virtual try-on. 

We further construct a human editing benchmark by adapting previous editing methods based on facial models to human body models (\textit{i.e.}, PTI~\cite{PTI} for image inversion, InterFaceGAN~\cite{shen2020interfacegan}, StyleSpace~\cite{wu2021stylespace}, and SeFa~\cite{sefa} for image manipulation). The impressive results in editing human clothes and attributes demonstrate the potential of the given model zoo in downstream tasks. In addition, a concurrent work, InsetGAN~\cite{fruhstuck2022insetgan}, is also evaluated with our baseline model, further showing the potential usage of our pre-trained human generative models.

Here is the summary of the main contributions of this paper: 
1) We collect a large-scale, high-quality, and diverse dataset, Stylish-Humans-HQ (SHHQ), containing $230K$ human full-body images for unconditional human generation task.
2) We investigate three crucial questions that have aroused broad interest in the community and discuss our observation through comprehensive analysis.
3) We build a model zoo for unconditional human generation to facilitate future research. An editing benchmark is also established to demonstrate the potential of the proposed model zoo.

%% file: section/related_work.tex
\section{Related Work}
\label{sec:related}

\subsection{Dataset For Human Generation}
\label{sec:related_dataset}

Large-scale and high-quality clothed human-centric training datasets are the critical fuel for the training of StyleGAN models. 
A qualified dataset should conform to the following aspects: 
\textbf{1)} \textit{Image quality}: high-resolution images with rich textures offer more raw detailed semantic information to the model. 
\textbf{2)} \textit{Data volume}: the size of dataset should be sufficient to avoid generative overfitting~\cite{arjovsky2017towards,zhang2019progressive}.  
\textbf{3)} \textit{Data coverage}: the dataset should cover multiple attribute dimensions to guarantee diversity of the model, for instance, gender, clothing type, clothing texture, and human pose. 
\textbf{4)} \textit{Data content}: since this report only focuses on the generation of single full-body human, occlusion caused by other people or objects is not considered here, whereas self-occlusion is taken into account. That is, each image should contain only one complete human body.

Publicly available datasets built particularly for full human-body generation are rare, but there are several practices~\cite{ma2017pose,Ma_2018_CVPR,Siarohin_2018_CVPR} cooperating with DeepFashion~\cite{liu2016deepfashion} and Market1501~\cite{market1501}. DeepFashion dataset~\cite{liu2016deepfashion} with well-labeled attributes and diverse garment categories is satisfactory for image classification and attribute prediction, but not adequate for unconditional human generation since it emphasizes fashion items rather than human bodies. Thus the number of close-up shots of clothing is much higher than that of full-body images. Market1501 dataset~\cite{market1501} fails for human generation tasks due to its low resolution ($128\times64$). 
There are some human-related datasets in other domains rather than GAN-based applications: 
datasets related to human parsing~\cite{Gong_2017_CVPR,atr} are limited by scalability and diversity; common datasets for virtual try-on tasks either contain only the upper body~\cite{han2018viton} or are not public~\cite{yildirim2019generating}. A detailed comparison of the above datasets in terms of data scale, average resolution, attributes labeling, and proportion of full-body images across the whole dataset 
is listed in Table~\ref{tab:comp1}. In general, there is no high-quality and large-scale full human-body dataset publicly available for the generative purpose.

\begin{table*}[t]
\begin{center}
  \caption{\textbf{Comparison of SHHQ with other publicly available datasets.} The proposed Stylish-Humans-HQ (SHHQ) dataset covers the largest number of human images to date and it is the only dataset that satisfies all four data requirements mentioned in Section~\ref{sec:related_dataset}. For each dataset, we report the total number of images, resolutions, label properties, and the proportion of full-body images in the dataset. ``Labeled Attributes'' records whether the dataset provides human-relevant labels, while ``Full-Body Ratio'' indicates the proportion of full-body images each dataset contains. 
  }
  \label{tab:comp1}
  \begin{tabular}{lcccc}
    \toprule
    Dataset & Total Image \# &  Mean Resolution & Labeled Attributes & Full-Body Ratio \\
    \midrule
    {ATR~\cite{atr}}                     & $7,700$  & $400\times600$  & \cmark & $76\%$\\    
    Mark1et1501~\cite{market1501}        & $32,668$ & $128\times64$ & \cmark & $100\%$ \\
    DeepFashion~\cite{liu2016deepfashion}& \specialcell{$146,680$}   & $1101\times750$ & \cmark & $6.8\%$\\
    LIP~\cite{Gong_2017_CVPR}            & $50,462$ & $196\times345$  & \cmark & \specialcell{$37\%$} \\    
    VITON~\cite{han2018viton}            & $16,253$ & $256\times192$ & \xmark & $0\%$\\
    \toprule
    \textbf{Stylish-Humans-HQ}                       & \textbf{231,176}  & \specialcell{$\mathbf{1024\times512}$ \\\textbf{up to $\mathbf{2240\times1920}$ in raw data}} &  \textbf{\cmark}  & $\mathbf{100\%}$\\
    \bottomrule
  \end{tabular}
\end{center}
\end{table*}

\subsection{StyleGAN}
In recent years, the research focus has gradually shifted to generating high-fidelity and high-resolution images through Generative Adversarial Networks~\cite{biggan,pggan}. The StyleGAN generator~\cite{stylegan} was introduced and became the state-of-the-art network of unconditional image generation. Compared to previous GAN-based architectures~\cite{wgan,wgan-gp,miyato2018spectralnorm}, SytleGAN injects a separate attribute factor (i.e., style) into the generator to influence the appearance of generated images. Then StyleGAN2~\cite{stylegan2} redesigns the normalization, multi-scale scheme, and regularization method to rectify the artifacts in StyleGAN images. The latest update to StyleGAN~\cite{stylegan3} reveals the non-ideal case of detailed textures sticking to fixed pixel locations and proposes an alias-free network. 

\subsection{Human Generation}
In human generation research, most of the existing applications focus on precise control of pose and appearance by leveraging conditional VAE and U-Net~\cite{esser2018variationalUNet,sarkar2021humangan} or StyleGAN-related architectures~\cite{posewithstyle,tryongan,grigorev2021stylepeople,styleposegan}. Specifically, the 3D method~\cite{grigorev2021stylepeople} renders StyleGAN-generated neural textures on the parametric human models, but the results are restricted by the quantity and quality of training data. The other works~\cite{sarkar2021humangan,styleposegan,posewithstyle} preserve texture quality by spatial modulation using the extracted UV texture map, and perform pose transfer conditioned by extracted pose features. The limitation of these works is that paired data with satisfied volume is required for training. Moreover, studies~\cite{esser2018variationalUNet,sarkar2021humangan,styleposegan,posewithstyle} trained with DeepFashion indicate that DeepFashion can produce decent results in human generation, at least for head-to-waist images. These works rely on additional network modifications and certain human priors. The above works can be summarized as ``network engineering''; they require some architectural changes and certain priors. In contrast, this study probes unconditional human generation challenges from a data perspective.

\subsection{Image Editing}
Benefiting from StyleGAN, one of the significant downstream applications is image editing~\cite{styleclip,shen2020interfacegan,wu2021stylespace,abdal2020image2styleganpp,xu2022transeditor}. A standard image editing pipeline usually involves inversion from a real image to the latent space and manipulating the embedded latent code. Existing works for \textit{image inversion} can be categorized into optimization-based~\cite{abdal2021styleflow,tewari2020pie}, encoder-based~\cite{PTI,e4e,Highfidelitygan}, and hybrid methods~\cite{chai2021ensembling}, which exploit encoders to embed images into latent space first and then refine with optimization. As for \textit{image manipulation}, studies explore the capability of attribute disentanglement in the latent space with supervised~\cite{abdal2021styleflow,shen2020interfacegan,jiang2021talk} and unsupervised~\cite{harkonen2020ganspace,sefa,tzelepis2021warpedganspace,wu2021stylespace} networks. In specific, 
Jiang~\emph{et al.}~\cite{jiang2021talk} proposes to use manually labeled fine-grained annotations to find non-linear manipulation directions in the StyleGAN latent space, while SeFa~\cite{sefa} search for semantic directions without supervision. StyleSpace~\cite{wu2021stylespace} defines the style space $S$ and proves that it is more disentangled than $W$ and $W+$ space. In this report, we perform image editing on real images by the inversion method PTI~\cite{PTI} and various directions through the chosen methods~\cite{shen2020interfacegan,sefa,wu2021stylespace} on our model to verify whether our model with human images could preserve the characteristics demonstrated on rigid objects. 

%% file: section/dataset.tex
\section{Stylish-Humans-HQ Dataset}
\label{sec:Preliminary}

To investigate the key factors in unconditional human generation task from a data-centric perspective, we propose a large-scale, high-quality, and diverse dataset, Stylish-Humans-HQ (SHHQ). In this section, we first present the data collection and preprocessing (Section~\ref{sec:data_col_pre}) , in which we construct the SHHQ dataset. Then, we analyze the data statistic (Section~\ref{sec:datastats}) of Stylish-Humans-HQ dataset to demonstrate the superiority of SHHQ compared to other datasets from a statistical perspective.

\begin{figure}[t]
  \centering
  \includegraphics[width=\linewidth]{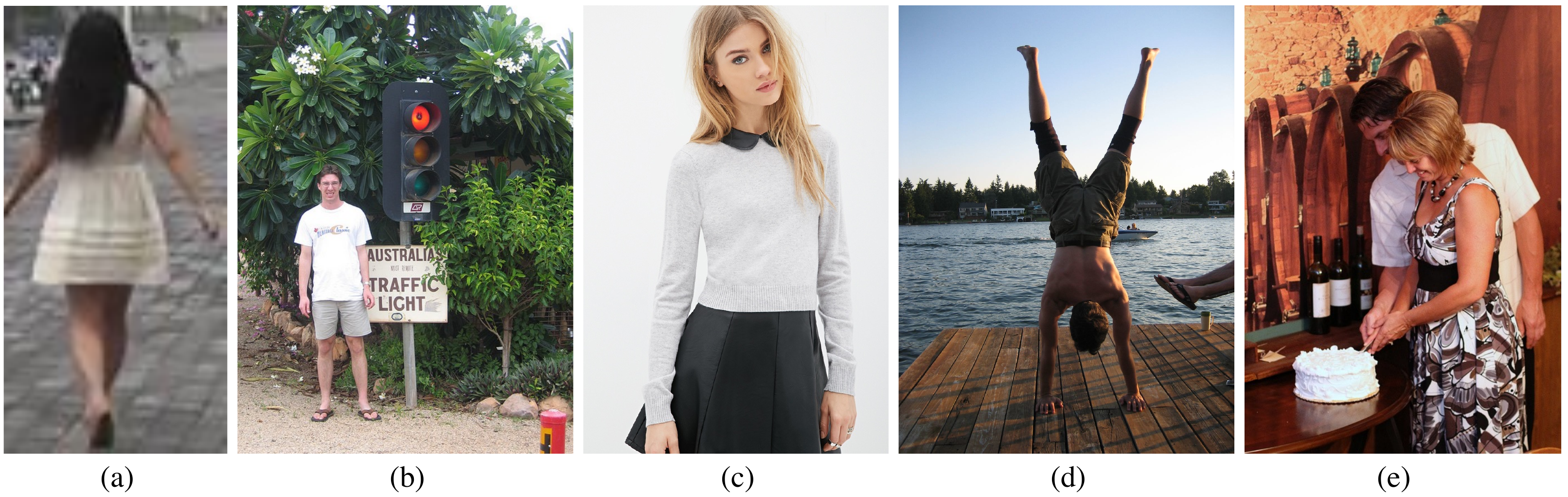}
  \caption{\textbf{Data Preprocessing.} The following types of images will be removed during our data preprocessing pipeline. (a) Images of low resolution. (b) Images of a person not placed in the center. (c) Images with missing body parts, \eg, with presence of the upper body only. (d) Extreme posture, \eg, handstand. (e) Images with multi-person. 
  }
  \label{fig:standardization}
\end{figure}

\begin{figure*}[t]
  \centering
  \includegraphics[width=\linewidth]{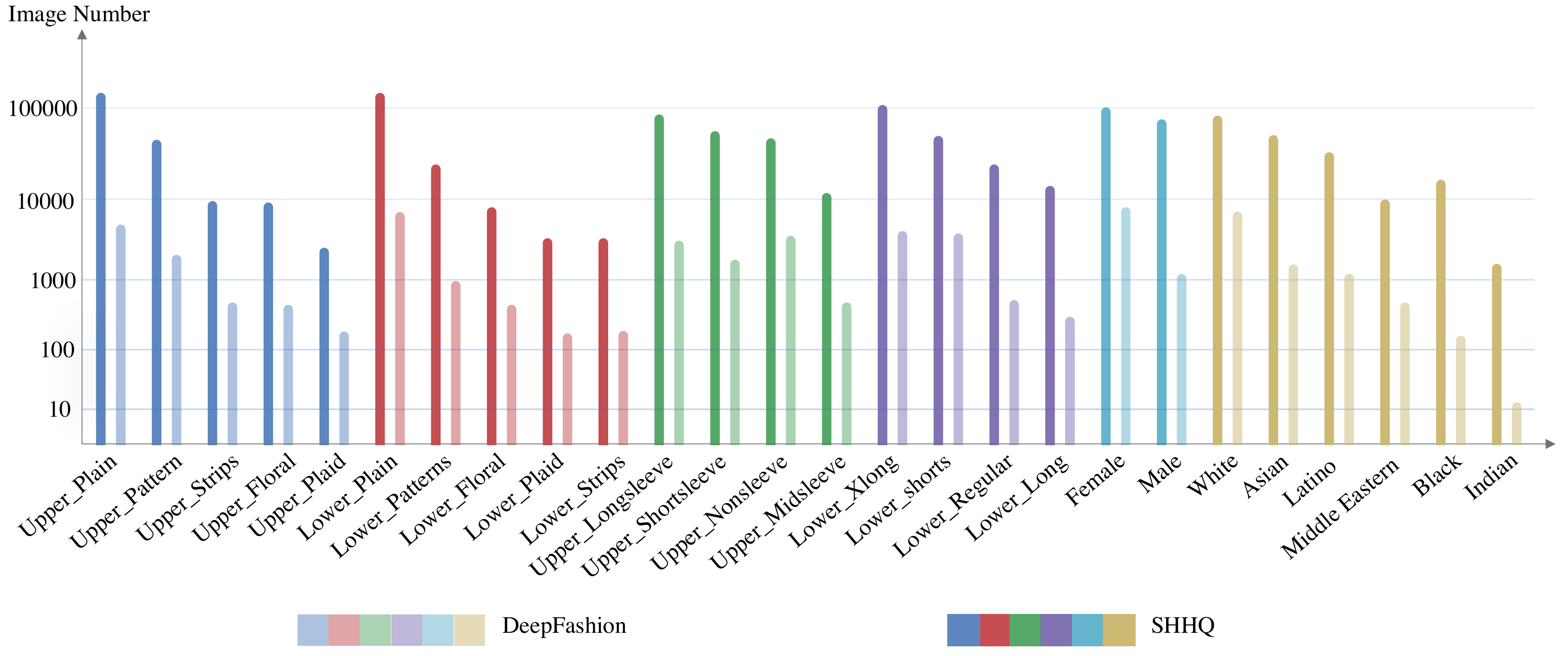}
  \caption{\textbf{Attribute Distribution.} Comparison of different attributes between the pruned DeepFashion and our Stylish-Humans-HQ dataset: Texture of the upper clothing, texture of the lower clothing, length of upper clothing, length of lower clothing, gender, and ethnicity. Note that the scale of y-axis uses \textit{logarithmic} scaling, with a base of 10.}  
  \label{fig:Attributes}
\end{figure*}

\subsection{Data Collection and Preprocessing} 
\label{sec:data_col_pre}
We first obtain over $500K$ raw data of human images from the Internet, covering a wide variety of races, ages and clothing styles. Some representative samples of raw data are shown in Appendix~\ref{sec:appendix_dataset}. We preprocess the data with six factors taken into consideration (\ie, resolution~\cite{liu2016deepfashion}, body position~\cite{stylegan}, body-part occlusion, human pose~\cite{liu2016deepfashion,stylegan}, multi-person, and background), which are critical for the quality of a human dataset. After the data preprocessing procedure, we obtain a clean dataset of $231,176$ images with high quality; see Figure~\ref{fig:align_vis} (a) for examples.

\noindent
\textbf{Resolution.} 
We discard images with a resolution lower than $1024\times512$ (Figure~\ref{fig:standardization} (a)).

\noindent
\textbf{Body Position.} The position of the body varies widely in different images, as shown in Figure~\ref{fig:standardization} (b). We design a procedure in which each person is appropriately cropped based on human segmentation~\cite{mmseg2020}, padded and resized to the same scale, and then placed in the image such that the body center is aligned. The body center is defined as the average coordinate of the entire body using segmentation.

\noindent
\textbf{Body-Part Occlusion.} This work aims at generating full-body human images, images with any missing body parts are removed (\eg, the half-body portrait shown in Figure~\ref{fig:standardization} (c)).

\noindent
\textbf{Human Pose.} 
We remove images with extreme poses (\eg, lying postures, handstand in Figure~\ref{fig:standardization} (d)) to ensure learnability of the data distribution. We exploit human pose estimation~\cite{openpose} to detect those extreme poses.

\noindent
\textbf{Multi-Person Images.} Some raw images contain multiple persons, such as in Figure~\ref{fig:standardization} (e). In this work, our goal is to generate single-person full-body images, so we keep unoccluded single-person full-body images, and remove those with occluded people.

\noindent
\textbf{Background.} Some images contain complicated backgrounds, requiring additional representation ability. To focus on the generation of the human body itself and eliminate the influence of various backgrounds, we use a segmentation mask~\cite{mmseg2020} to modify the image background to pure white. The edges of the mask are smoothed by Gaussian blur.

\subsection{Data Statistics}
\label{sec:datastats}

Table~\ref{tab:comp1} presents the comparison between SHHQ and other public datasets.
\noindent
\textbf{Dataset Scale.} As shown in the table, our proposed SHHQ is currently the largest dataset in scale compared to other datasets. Among them, the data volume of the SHHQ dataset is $1.6$ times that of DeepFashion~\cite{liu2016deepfashion} dataset and is much larger than that of others ($30$ times to ATR~\cite{atr}, $7$ times to Market1501~\cite{market1501}, $4.6$ times to LIP~\cite{Gong_2017_CVPR}, and $14$ times to VITON~\cite{han2018viton}). 
\noindent
\textbf{Resolution.} Images from ATR~\cite{atr}, Market1501~\cite{market1501}, LIP~\cite{Gong_2017_CVPR}, and VITON~\cite{han2018viton} are lower in resolution, which is insufficient for our generation task, while the proposed SHHQ and DeepFashion provide high-definition images up to $2240\times1920$. 
\noindent
\textbf{Labels.} All datasets beside VITON provide various labeled attributes. Specifically, DeepFashion~\cite{liu2016deepfashion}  and SHHQ label the clothing types and textures, which is useful for human generation/editing tasks. Full-body ratio denotes the proportion of full-body images in the dataset. 
\noindent
\textbf{Full-Body Ratio.} Although DeepFashion~\cite{liu2016deepfashion}  offers over $146K$ images with decent resolution, only $6.8\%$ of them are full-body images, while SHHQ achieves a $100\%$ full-body ratio. Appendix~\ref{sec:appendix_dataset} shows a visual comparison among several representative datasets and the proposed SHHQ dataset.

In summary, SHHQ covers the largest number of human images with high-resolution, labeled clothing attributes, and $100\%$ full-body ratio. It again confirms that our dataset is more suitable for full-body human generation than other public datasets.

Of all the datasets compared above, DeepFashion~\cite{liu2016deepfashion} is the most relevant to our human generation task. In Figure~\ref{fig:Attributes}, we further present the comparison of different attributes between filtered DeepFashion~\cite{liu2016deepfashion}  (removing occluded body) and SHHQ in a more detailed view. The bar chart depicts the distributions along six dimensions: upper cloth texture, lower cloth texture, upper cloth length, lower cloth length, gender, and ethnicity. In particular, the number of females is approximately $4$ times the number of males in filtered DeepFashion~\cite{liu2016deepfashion} , while our dataset features a more balanced female-to-male ratio of $1.49$. With the help of DeepFace API~\cite{serengil2021lightface}, it is shown that SHHQ is more diverse in terms of ethnicity. Advantages are also shown in the other five attributes. In terms of garment-related attributes, images with specific labels in filtered DeepFashion~\cite{liu2016deepfashion}  are too scarce to be used as a training set. The Stylish-Humans-HQ dataset boosts the number of each category by an average of $24.4$ times.

%% file: section/investigation.tex
\section{Systematic Investigation}

Our investigations are built on the official StyleGAN2 codebase~\footnote{\url{https://github.com/NVlabs/stylegan2-ada-pytorch}} and StyleGAN2 architecture. The detailed training settings can be found in Appendix~\ref{sec:appendix_training_scheme}. 

We conduct extensive experiments to study three factors concerning the quality of generated images: 1) data size (Section~\ref{sec:datasize}), 2) data distribution (Section~\ref{sec:data_distribution}), and 3) data alignment (Section~\ref{sec:Alignment}).

\begin{figure*}[t]
  \centering
  \includegraphics[width=\linewidth]{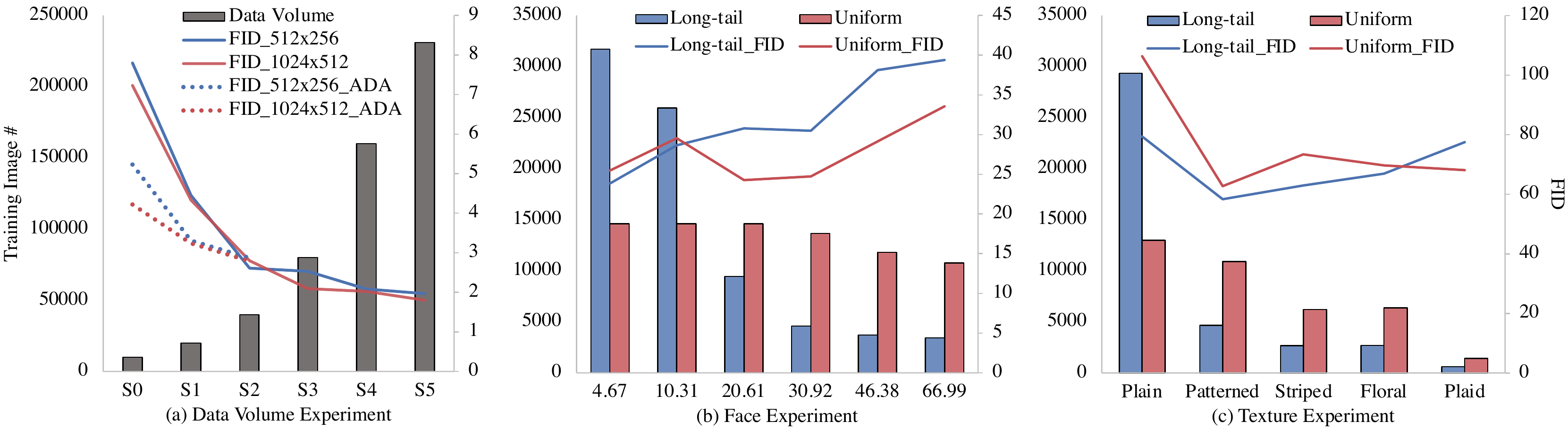}
  \caption{\textbf{Experiment results.} (a) FID scores for experiments $S0$ - $S5$ in $1024\times512$ and $512\times256$ resolutions. Dotted lines shows the FID scores of the models trained with ADA. (b) Bin-wise FIDs of long-tailed and uniform distribution in terms of facial yaw angle along with the number of training images (c) Bin-wise texture FIDs of long-tailed and uniform distribution along with the number of training images.}
  \label{fig:experimentcombine}
\end{figure*}

\subsection{Data Size}
\label{sec:datasize}
\noindent\textbf{Motivation.} 
Data size is an essential factor that determines the quality of generated images. Previous literature always takes different strategies to improve the generation performance according to different dataset sizes: regularization techniques~\cite{biggan} are employed to train a large dataset, while augmentation~\cite{stylegan2ada,toutouh2020data,zhao2020differentiable} and conditional feature transferring~\cite{bagan,wu2020conditional} are proposed to tackle the limited data of faces and non-rigid objects. Here, we design sets of experiments to examine the relationship between training data size and the image quality of generated humans.

\noindent\textbf{Experimental Settings.} 
To determine the relationship between data size and image quality for the unconditional human GAN, we construct $6$ sub-datasets and denoted these subsets as $S0$ ($10K$), $S1$ ($20K$), $S2$ ($40K$), $S3$ ($80K$), $S4$ ($160K$) and $S5$ ($230K$). Here, $S0$ is the pruned DeepFashion dataset.
We perform the training on two resolution settings for each set: $1024\times512$ and $512\times256$. 
Considering the case of limited data, we also conduct additional training experiments with adaptive discriminator augmentation (ADA)~\cite{stylegan2ada} for small datasets $S0$, $S1$, and $S2$.
Fr\'echet Inception Distance (FID) and Inception Score (IS) are the indicators for evaluating the model performance.

\noindent\textbf{Results.}
As shown in Figure~\ref{fig:experimentcombine} (a), the FID scores (solid lines) decrease as the size of the training dataset increases for both resolution settings. The declining trend is gradually flattening and tends to converge. $S0$ generates the least satisfactory results, with FID of $7.80$ and $7.23$ for low- and high-resolution, respectively, while $S1$ achieves corresponding improvements of $42\%$ and $40\%$ on FID with only an additional $10K$ training images. When the training size reaches $40K$ for both resolutions, the FID curves start to converge to a certain extent.

The dotted lines indicate the results of ADA experiments with subsets $S0$ - $S2$. The employed data augmentation strategy helps to reduce FID when training data is less than $40K$.
Table~\ref{tab:fid_is} and Figure~\ref{fig:isscore} in Appendix~\ref{sec:appendix_expresults} show the detailed quantitative results of FID and IS scores, from where the IS increases with the size of training data and slows down after the amount of data reaches $40K$.

\noindent\textbf{Discussion.} 
The experiments confirm that ADA can improve the generation quality for datasets smaller than $40K$ images, in terms of FID and IS. However, ADA still cannot fully compensate for the impact of insufficient data. Besides, when the amount of data is less than $40K$, the relationship between image quality and data size is close to linear. As the amount of data increases to $40K$ and more, the improvement in the quality of the resulting images slows down and is less significant.

\subsection{Data Distribution}
\label{sec:data_distribution}

\noindent\textbf{Motivation.} 
The nature of GAN makes the model inherits the distribution of the training dataset and introduces generation bias due to dataset imbalance~\cite{liu2019large}. This bias severely affects the performance of GAN models. To address this issue, studies for unfairness mitigation~\cite{tailgan,biascorrection,bagan,fairnessgan,fairgan} have attracted substantial research interest. In this work, we explore the question of data distribution in human generation and conduct experiments to verify whether a uniform data distribution can improve the performance of a human generation model.

\begin{figure*}[t]
  \centering
  \includegraphics[width=\linewidth]{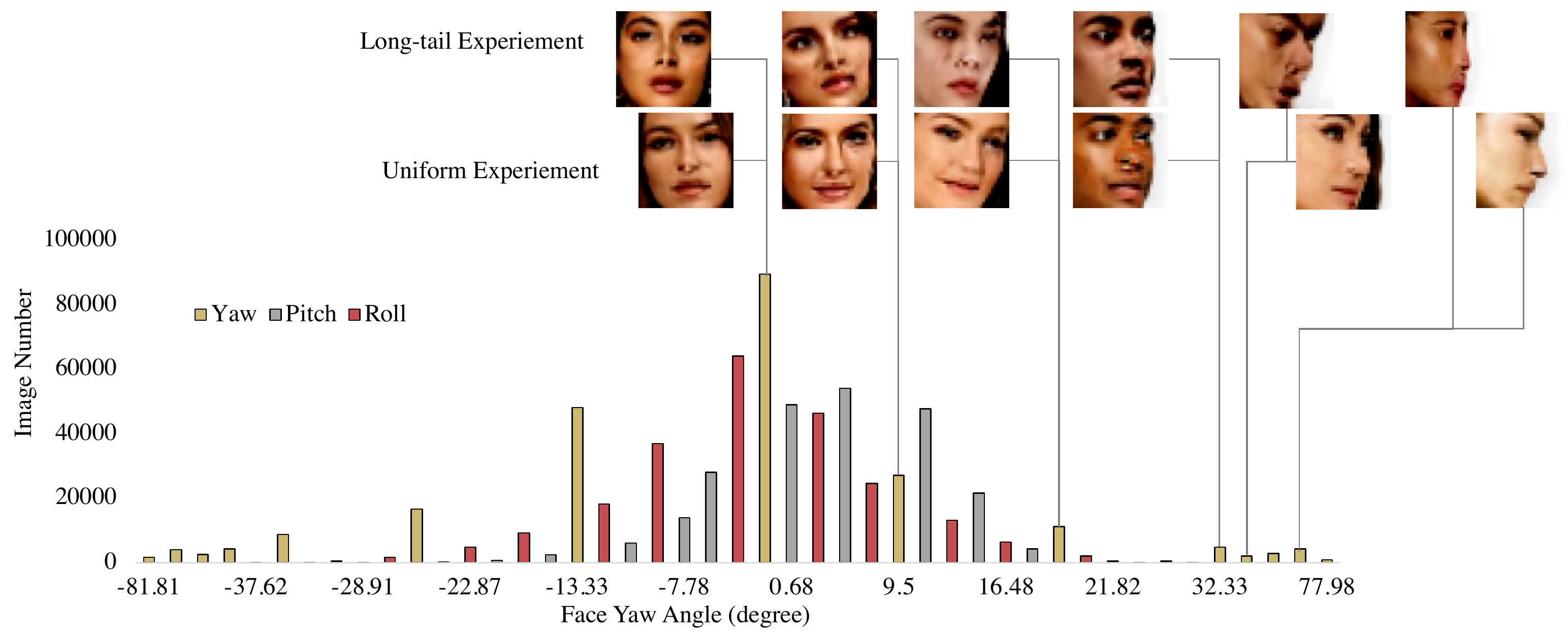}
  \caption{\textbf{Distribution of face orientation with face visualization.} The distribution of head rotation angles along the yaw/pitch/roll dimensions in the proposed SHHQ dataset, where yaw represents head rotation about the vertical axis. $0^\circ$ in yaw represents the face facing straight ahead. The face patches are cropped from the \textit{generated} images at $1024\times512$ resolution (see Section~\ref{sec:data_distribution}). Face patches in the first row are randomly sampled from the long-tailed experiment, and the second row is for the uniform experiment. Face yaw angle increases from left to right.}
  \label{fig:all_non_repeat_face_angle}
\end{figure*}

\noindent\textbf{Experimental Settings.} This study decomposes the distribution of the human body into Face Orientation and Clothing Texture, since face fidelity has a significant impact on visual perception and clothing occupies a large portion of the full-body image. Figure~\ref{fig:all_non_repeat_face_angle} depicts the distribution of face orientation angle in SHHQ. The general features of human faces are relatively symmetrical; thus, we fold yaw distribution vertically along $0^\circ$ and get the long-tailed distribution. For the face and clothing experiments, we collect an equal number of long-tailed and uniformly distributed datasets from SHHQ for face rotation angle and upper-body clothing texture, respectively.

\noindent\textbf{Results.}
To evaluate the image quality in terms of different distributions, the cropped faces and clothing regions are used to calculate FID, and FID is calculated separately for each bin. Result can be found in Figure~\ref{fig:experimentcombine} (b) and (c).

1) Face Orientation:
As for the long-tailed experiment (blue curve in Figure~\ref{fig:experimentcombine} (b)), the FID progressively grows as the face yaw angle increases and remains high when the facial rotation angle is too large. By contrast, the upward trend for the face FID in the uniform experiment (red) is more gradual. 
In addition, the amount of the training data of the first two bins in the uniform set is greatly reduced compared to the long-tail experiment, but the damage to FID is slight. 

Figure~\ref{fig:all_non_repeat_face_angle} presents the random samples belonging to different bins in face experiments. It can be visually observed that the right-most samples in the uniform experiment have better image quality. More cropped faces in different bins from both experiments can be found in Appendix~\ref{sec:appendix_expresults}.

2) Clothing Texture:
From Figure~\ref{fig:experimentcombine} (c), except for the first bin (``plain'' pattern), the FID curve climbs steadily as the amount of training data for the long-tailed experiment decreases, and the FID curve for the uniform experiment also shows a near-uniform pattern. In particular, FID of the last bin for the uniform experiment is lower than that in the long-tailed setting. We infer that the training samples for ``plaid'' clothing texture in the long-tailed experiment are too few to be learned by the model.  

As for the ``plain'' bin results, the long-tailed distribution has a lower FID score in this bin. The reason may lie in that the number of plain textures in the long-tailed distribution is considerably higher than that in the uniform distribution. Also, it can be observed that the training patches in this bin are mainly textureless color blocks (see Appendix~\ref{sec:appendix_expresults}), where  such patterns may be easier to capture by models.

\noindent\textbf{Discussion.} 
Based on the above analysis, we conclude that the uniform distribution of face rotation angles can effectively reduce the FID of rare training faces while maintaining acceptable image quality for the dominant faces. However, simply balancing the distribution of texture patterns does not always reduce the corresponding FID effectively. This phenomenon raises an interesting question that can be further explored: is the relation between image quality and data distribution also entangled with other factors, \eg, image pattern and data size?
Additionally, due to the nature of GAN-based structures, a GAN model memorizes the entire dataset, and usually, the discriminator tends to overfit those poorly sampled images at the tail of the distribution. Consequently, the long-tailed situation accumulated as ``tail'' images is barely generated. From this perspective, it also can be seen that the uniform distribution preserves the diversity of faces/textures and partially alleviates this problem.

\begin{figure*}
  \centering
  \includegraphics[width=\linewidth]{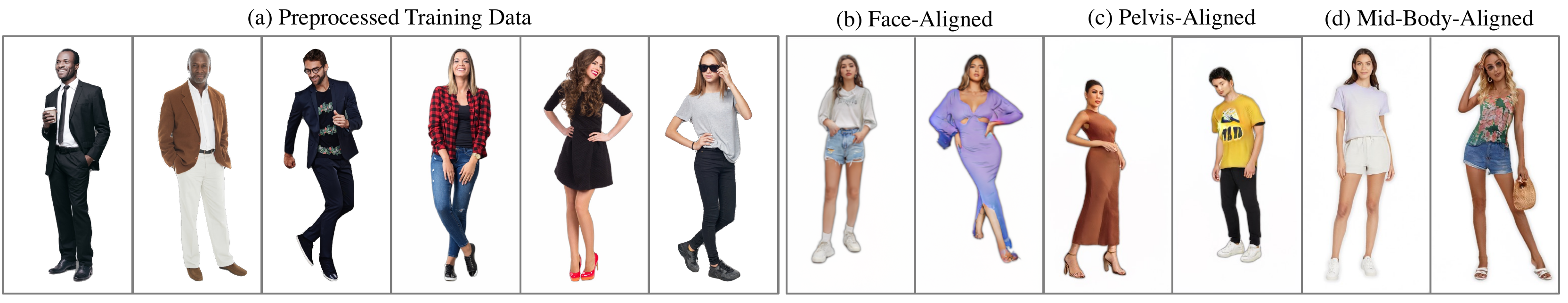}
  \caption{\textbf{Example of Preprocessed Data and Different Alignment Schemes.} Part (a) illustrates processed training data with the consideration of resolution, body position, body-part occlusion, human pose, multi-person and background. (b) - (d) display random sample results from baseline models with three different alignment strategies. (b) Face-aligned: aligned with averaged coordinate from a face bounding box. (c) Pelvis-aligned: the keypoint~\cite{openpose} of the pelvis is used as the alignment center. (d) Mid-body-aligned: body center is defined by averaging coordinates of extreme points (topmost, bottommost, leftmost, and rightmost points) on the boundaries of the segmentation mask. 
  }
  \label{fig:align_vis}
\end{figure*}

\subsection{Data Alignment}
\label{sec:Alignment}
\noindent\textbf{Motivation.} 
Recently, researchers have drawn attention to spatial bias in generation tasks. Several works~\cite{stylegan,kazemi2014one} align face images with keypoints for face generation, and other studies propose different alignment schemes to preprocess non-rigid objects~\cite{liu2016fashion,biggan,diffusiongan,styleganxl,jiang2021robust}. In this paper, we study the relationship between the spatial deviation of the entire human body and the generated image quality.

\noindent\textbf{Experimental Settings.}
We randomly sample a set of $50K$ images from the SHHQ dataset and align every image separately using three different alignment strategies: aligning the image based on the face center, pelvis, and the midpoint of the whole body, as shown in Figure~\ref{fig:align_vis}. 

Following are the reasons for selecting these three positions as alignment centers. 1) For the face center, we hypothesize that faces contain rich semantic information that is valuable for learning and may account for a heavy proportion in human generation. 
2) For the pelvis, studies related to human pose estimation~\cite{martinez2017simple,pavlakos2017coarse,veges2019absolute,nie2019single} conventionally predict the body joint coordinates relative to the pelvis. Thus we employ the pelvis as the alignment anchor. 
3) For the body's midpoint, the leg-to-body ratio (the proportion of upper and lower body length) may vary among different people; therefore, we try to find the mean coordinates of the full body with the help of the segmentation mask.

\noindent\textbf{Results.}
Human images are complex and easily affected by various extrinsic factors such as body poses and camera viewpoints. The FID scores for the face-aligned, pelvis-aligned, and mid-body-aligned experiments are $3.5$, $2.8$, and $2.4$, respectively. Figure~\ref{fig:align_vis} further interprets this perspective as the human bodies in (b) and (c) are tilted, and the overall image quality is degraded. The example shown in Figure~\ref{fig:align_vis} (c) also presents the inconsistent human positions caused by different leg-to-body ratios.


\begin{figure*}[h]
  \centering
  \vspace{5pt}
  \includegraphics[width=\linewidth]{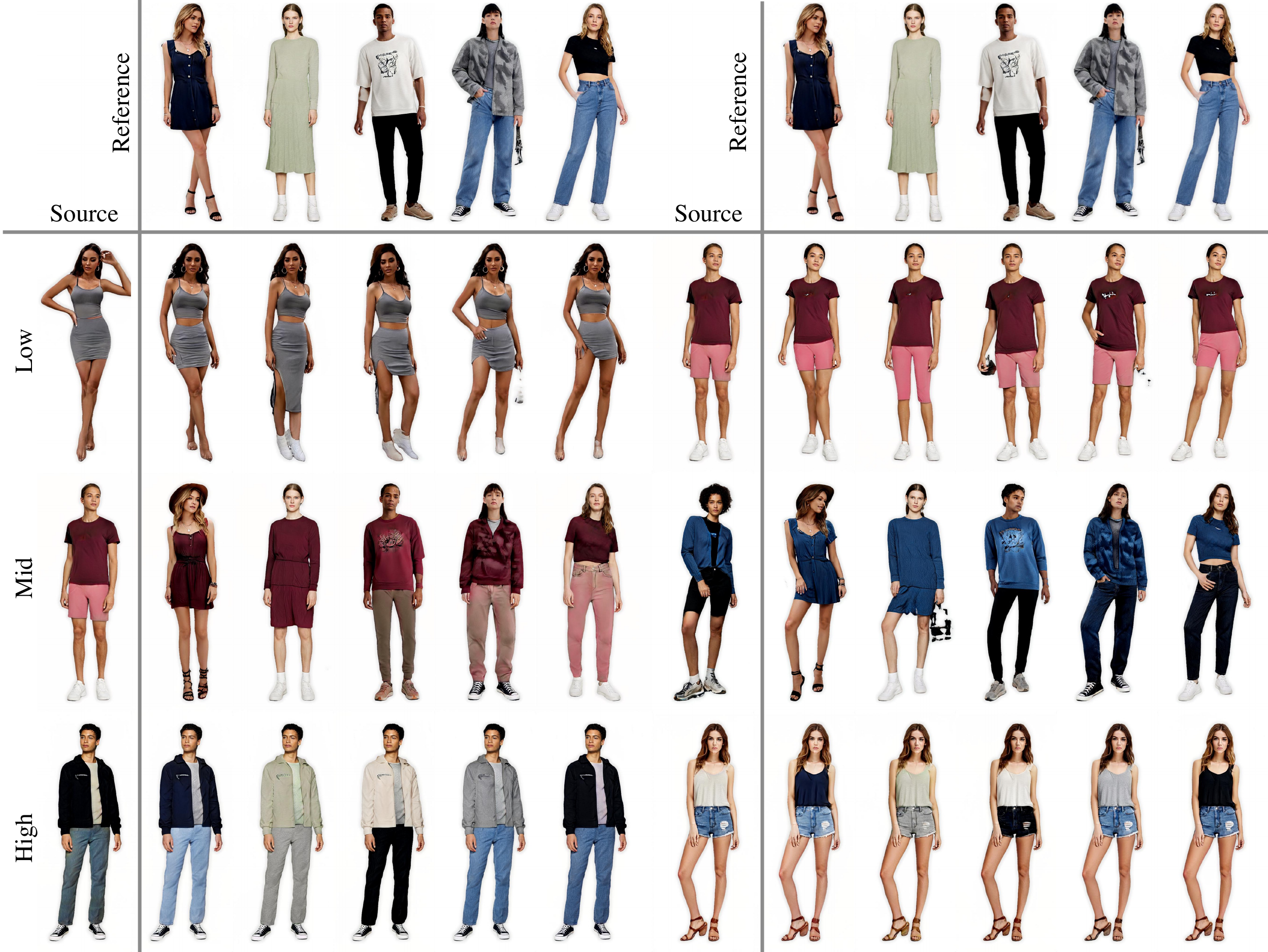}
  \caption{\textbf{Style-mixing Results.} The sets of images are randomly sampled from the provided baseline model, with latent codes recorded. Those images are treated as reference and source images. The rest of the images are generated by style-mixing: borrowing low/mid/high layers in the corresponding reference images' latent codes and combining them with the rest layers of latent code in source images. Low/mid/high corresponds to coarse/middle/fine spatial resolution. It can be seen that low layers in the latent code control coarse features such as poses, middle layers are related to clothing type, identical appearance, and higher layers convey fine-grained features, for example, clothing color.}
  \label{fig:stylemixing}
\end{figure*}

\begin{figure*}[h]
  \centering
  \includegraphics[width=\linewidth]{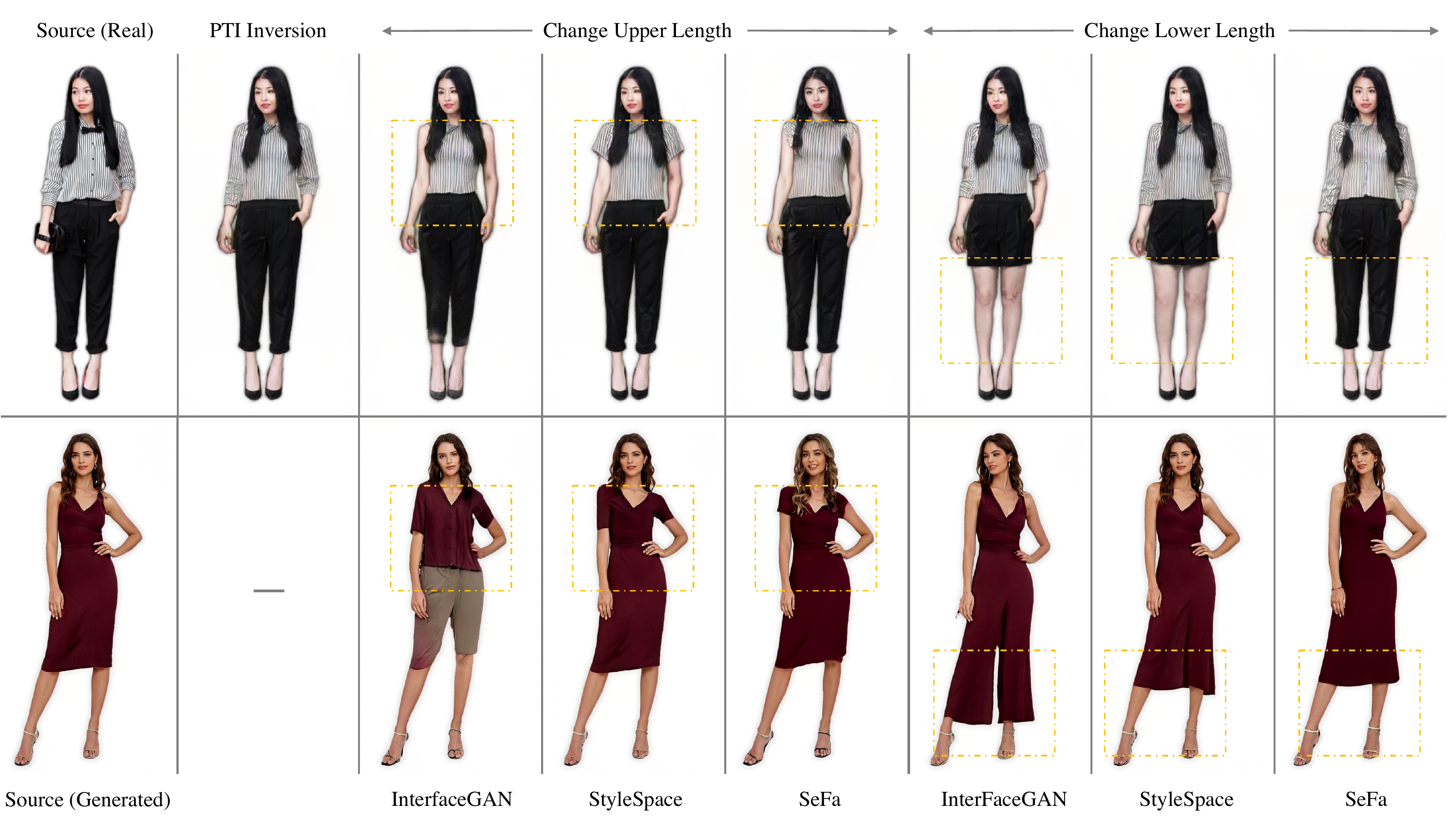}
  \caption{\textbf{Image editing results with different methods.} Image editing techniques are tested in both real images with PTI inversion (first row) and generated images from our baseline model (second row). The reference image is on the left, the second column shows the inverted result, and the remaining shows the results of modifying sleeve length and dress length. We shorten the length of upper and lower garments for the real image, and increase the upper/lower garment length for the generated image.}
  \label{fig:editingdemo}
\end{figure*}

\noindent\textbf{Discussion.} Both FID scores and visualizations suggest that the human generative models gain more stable spatial semantic information through the mid-body alignment method than face- and pelvis-centered methods. We believe this observation could benefit later studies on human generation. 

\subsection{Experimental Insights}
Now the questions can be answered based on the above investigations: 

\noindent For \textbf{Question-1} (Data Size): A large dataset with more than $40K$ images helps to train a high-fidelity unconditional human generation model, for both $512\times256$ and $1024\times512$ resolution.

\noindent For \textbf{Question-2} (Data Distribution): The uniform distribution of face rotation angles helps reduce the FID of rare faces while maintaining a reasonable quality of dominant faces. But simply balancing the clothing texture distribution does not effectively improve the generation quality.

\noindent For \textbf{Question-3} (Data Alignment): Aligning the human by the center of the full body presents a quality improvement over aligning the human by face or pelvis centers.

%% file: section/model_zoo.tex
\section{Model Zoo and Editing Benchmark}
\label{sec:baseline}

\subsection{Model Zoo}
In the field of face generation, a pre-trained  StyleGAN~\cite{stylegan} model has shown remarkable potential and success in various downstream tasks, including editing~\cite{abdal2021styleflow,wu2021stylespace}, neural rendering~\cite{gu2021stylenerf}, and super-resolution~\cite{chan2021glean,menon2020pulse}, which have spawned a series of compelling research.
Nevertheless, a publicly available pre-trained model is still lacking for the human generation task. To fill this gap, we train our baseline model on the collected $230K$ images (SHHQ) using the StyleGAN2~\cite{stylegan2} framework. The training takes about $5$ days on $8$ Tesla V100 GPUs, and the model provides the best FID of $1.57$. As seen in Figures~\ref{fig:generatedsample} and \ref{fig:concat_samples}, our model has the ability to generate full-body images with diverse poses and clothing textures under satisfactory image quality. 
The other two models are fully trained with StyleGAN~\cite{stylegan} and StyleGAN3~\cite{stylegan3}, both in a resolution of $1024\times512$. To adapt various application scenarios, we train models with different StyleGAN architectures~\cite{stylegan,stylegan2,stylegan3} in a lower image resolution ($512\times256$) as well. In total, all the $6$ models will be released for future research. We believe they can contribute to the exploration of various tasks related to human generation and continuously benefit the community.

Furthermore, the style mixing results of the baseline model show the interpretability of the corresponding latent space, which can be seen in Figure~\ref{fig:stylemixing}. As seen in the figure, source and reference images are sampled from the baseline model, and the rest images are the style-mixing results. We see that copying low layers from reference images to source images brings changes in geometry features (pose) from reference to the source. In contrast, other features such as skin colors, garment colors, and personal identities in source images are preserved. When copying middle styles, clothing type and identical appearance are copied from reference to the source. Finally, we observe that fine styles from high-resolution layers control the clothing color. More examples are displayed in Appendix~\ref{sec:appendix_stylemixing}. These style mixing results suggest that the provided model's geometry and appearance information are well disentangled.

\subsection{Editing Benchmark}

StyleGAN has presented remarkable editing capabilities over faces. In this section, we extend it to the full-scale human by using off-the-shelf inversion and editing methods, in which we validate the potential of our proposed model zoo. In addition, we re-implement the concurrent human generation method, InsetGAN~\cite{fruhstuck2022insetgan}, to further demonstrate another practical usage with the provided model zoo.

First, we leverage several SOTA StyleGAN-based facial editing techniques, such as InterFaceGAN~\cite{shen2020interfacegan}, StyleSpace~\cite{wu2021stylespace}, and SeFa~\cite{sefa}, with multiple editing directions: garment length for tops and bottoms, and global pose orientation. To examine the ability of editing real images with the provided model, PTI~\cite{PTI} is adopted to invert images before editing.

As illustrated in Figure~\ref{fig:editingdemo}, PTI presents the ability to invert real full-body human images. For attributes manipulation, StyleSpace~\cite{wu2021stylespace} expresses better disentanglement compared to InterFaceGAN~\cite{shen2020interfacegan} and SeFa~\cite{sefa}, as only the attribute-targeted region has been changed. However, as for the regions to be edited, the results of InterFaceGAN~\cite{shen2020interfacegan} are more natural and photo-realistic. It turns out that the latent space of the human body is more complicated than other domains such as faces, objects, and scenes, and more attention should be paid to disentangle human attributes. More editing results are shown in Appendix~\ref{sec:appendix_editing}.

\begin{figure*}[h]
  \centering
  \vspace{5pt}
  \includegraphics[width=\linewidth]{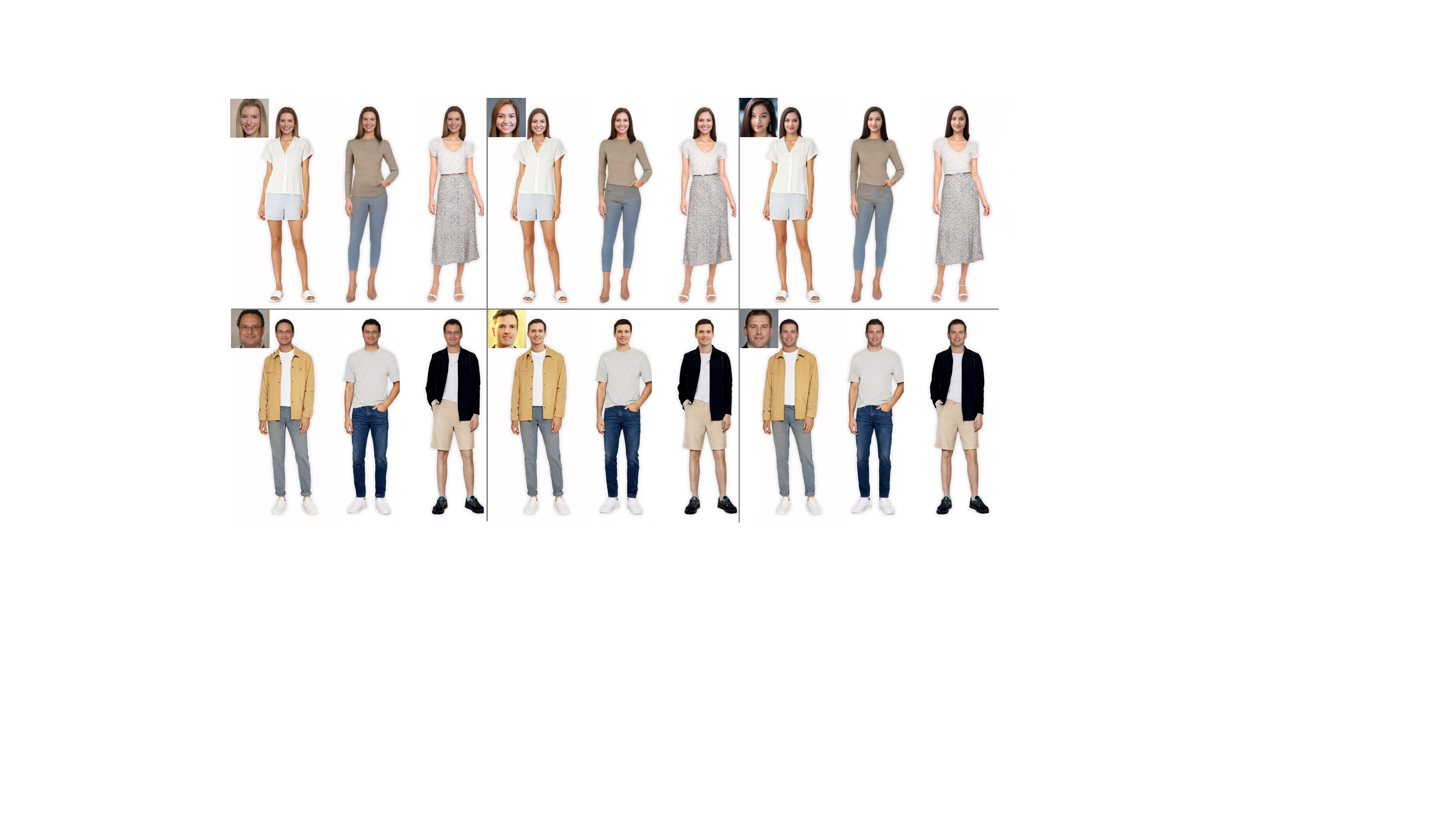}
  \caption{\textbf{InsetGAN Results.} We show the combined results of six different human bodies generated from the given baseline model and six faces generated from the FFHQ~\cite{stylegan2} model. For every single face, shown on the left corner of each grid, we jointly optimize it with three different bodies. 
}
  \label{fig:insetgan}
\end{figure*}

Moreover, InsetGAN~\cite{fruhstuck2022insetgan} proposes a multi-GAN optimization method to fuse face and body generated from separate GAN models. We re-implement this process by iteratively optimizing the latent codes for random faces and bodies generated by the FFHQ~\cite{stylegan2} and our baseline model, respectively. In Figure~\ref{fig:insetgan}, we show the fused full-body images of six human postures with different male and female faces. The optimization procedure blends diverse faces and bodies in a graceful manner. Our baseline model can be adapted to different faces and generate more complicated and diverse full-body images.

In this section, we adopt a representative facial editing method on humans generated by our pre-trained model and obtain impressive results. We also show that images generated from the released human model can be further locally optimized using existing GAN models. All these works demonstrate the effectiveness and convenience of our provided model zoo and verify its potential in human-centric tasks.

%% file: section/future_work.tex
\section{Future Work}
\label{sec:future_work}

In this study, we take a preliminary step towards the exploration of the human generation/editing problem. We believe many future works can be further explored based on the SHHQ dataset and the provided model zoo. In the following, we discuss three interesting directions, \ie, Human Generation/Editing, Neural Rendering, and Multi-modal Generation.

\noindent\textbf{Human Generation / Editing.} Studies in unconditional human generation~\cite{fruhstuck2022insetgan,yildirim2019generating}, human editing~\cite{posewithstyle,grigorev2021stylepeople,styleposegan}, virtual try-on~\cite{tryongan,Multi-Pose_tryon,unpair_tryon,Wang_2018_ECCV}, and motion transfer~\cite{chan2019everybody,liquidwarpinggan} heavily rely on large datasets to train or use existing pre-trained models as the first step of transfer learning. Furthermore, editing benchmarks show that disentangled editing of the human body remains challenging for existing methods~\cite{shen2020interfacegan, wu2021stylespace}. In this context, the released model zoo could expedite such research progress. Additionally, we further analyze failure cases generated by the provided model and discuss corresponding potential efforts that could be made to human generation tasks in Appendix~\ref{sec:appendix_limitations}.

\noindent\textbf{Neural Rendering.} Another future research direction is to improve 3D consistency and mitigate artifacts in full-body human generation through neural rendering~\cite{schwarz2020graf,niemeyer2021giraffe,eg3d,chan2021pigan,gu2021stylenerf,or2021stylesdf}. Similar to work such as EG3D~\cite{eg3d}, StyleNeRF~\cite{gu2021stylenerf}, and StyleSDF~\cite{or2021stylesdf}, we encourage researchers to use our human models to facilitate human generation with multi-view consistency.

\noindent\textbf{Multi-modal Generation.} Cross-modal representation is an emerging research trend, such as CLIP~\cite{radford2021clip} and Imagebert~\cite{qi2020imagebert}. Hundreds of studies are made on text-driven image generation and manipulation~\cite{jiang2021talk,xu2018attngan,dalle,qi2020imagebert,styleclip,wang2021faces,xia2021tedigan,kocasari2022stylemc}, \eg, DALLE~\cite{dalle} and AttnGAN~\cite{xu2018attngan}. In the meantime, several studies show interest in probing the transfer learning benefits of large-scale pre-trained models~\cite{radford2021clip,wu2020textgail,ghadiyaram2019large}. Most of these works focus on faces and objects, whereas research fields related to full-scale humans could be explored more, for example, text-to-human generation and text-driven human attributes manipulation, with the help of the provided full-body human models.

%% file: section/conclusion.tex
\section{Conclusion}
\label{sec:conclusions}

This work mainly probes how to train unconditional human-based GAN models to generate photo-realistic images from a data-centric perspective. By leveraging the $230K$ SHHQ dataset, we analyze three fundamental yet critical issues that the community cares most about: data size, data distribution, and data alignment. While experimenting with StyleGAN and large-scale data, we obtain several empirical insights. Apart from these, we create a model zoo, consisting of six human-GAN models, and the effectiveness of the model zoo is demonstrated by employing several state-of-the-art face editing methods.

\noindent
\textbf{Acknowledgements.} We thank Hao Zhu, Zhaoyang Liu and Zhuoqian Yang for their feedback and discussions. This study is partly supported by NTU NAP, MOE AcRF Tier 1 (2021-T1-001-088), and under the RIE2020 Industry Alignment Fund Industry Collaboration Projects (IAF-ICP) Funding Initiative, as well as cash and in-kind contribution from the industry partner(s).

%% file: section/appendix.tex
\appendix

\noindent
\textbf{\LARGE Appendix}

\section{SHHQ: StyleGAN-Human Datasets}
\label{sec:appendix_dataset}
\begin{figure*}[h]
  \centering
  \includegraphics[width=\textwidth]{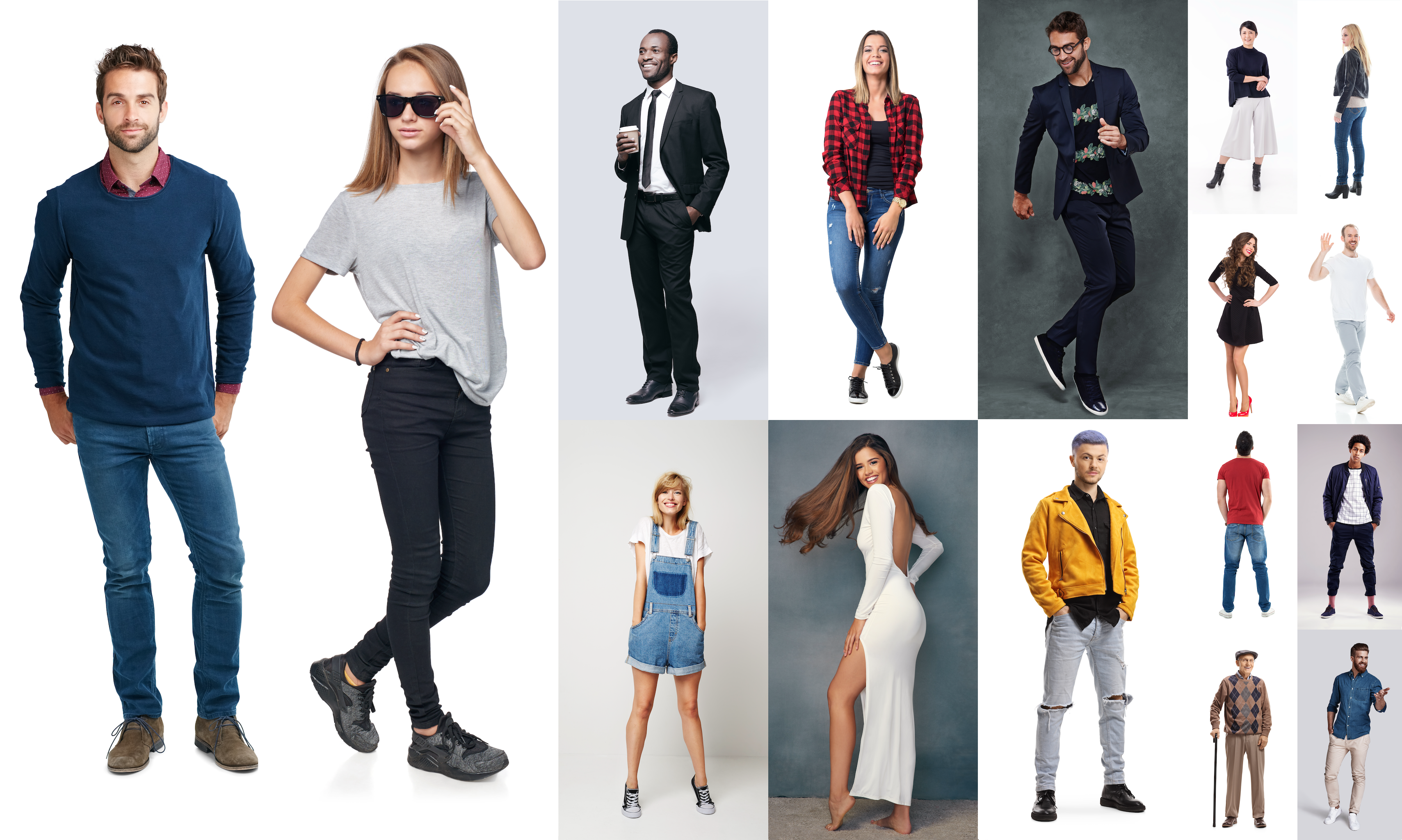}
  \caption{\textbf{Examples of raw data in the training dataset.}}
  \label{fig:trainingsamples}
\end{figure*}

\begin{figure}
  \centering
  \includegraphics[width=0.8\linewidth]{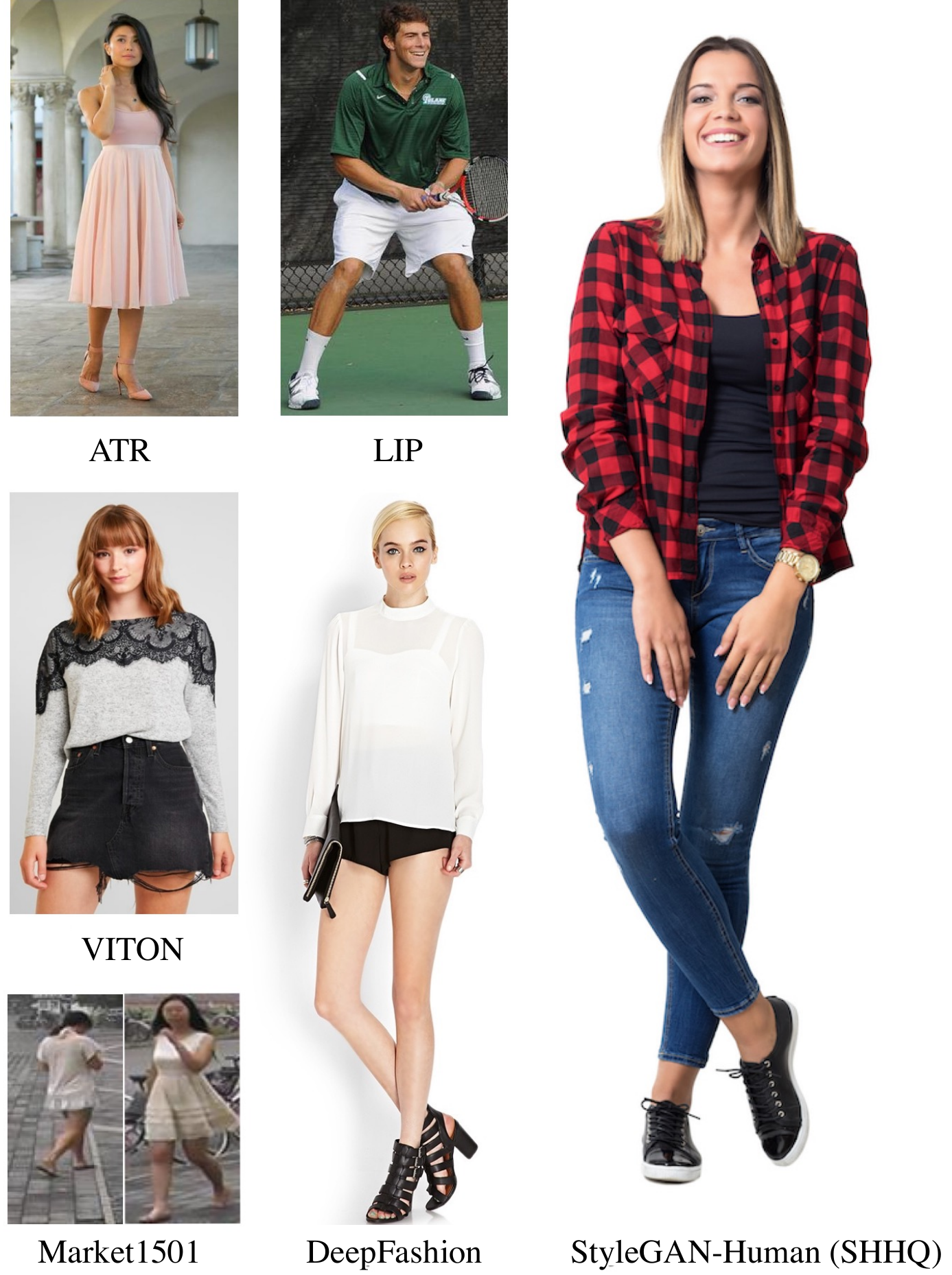}
  \caption{\textbf{Samples from different dataset with diverse resolution.}
  }
  \label{fig:dataset_comp}
\end{figure}

\begin{figure*}
  \centering
  \includegraphics[width=0.9\textwidth]{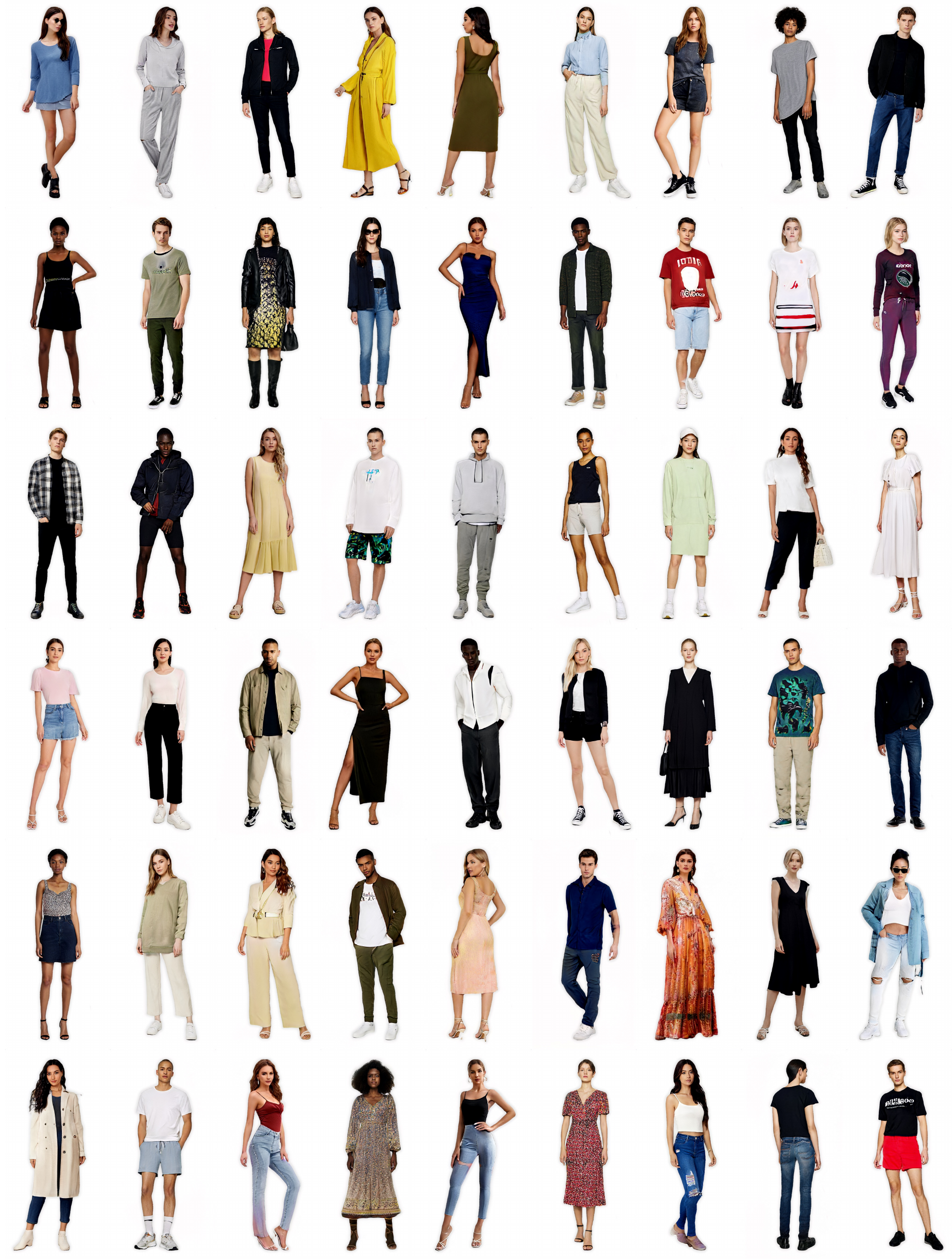}
  \caption{\textbf{Samples from our baseline model.} The model has shown the ability of generating random person with diverse clothing types, poses, genders, races, and hair types.}
  \label{fig:concat_samples}
\end{figure*}

The dataset we collected consists of $230K$ high-quality images of humans that vary in clothing appearance, ethnicity, and pose. Several training samples are shown in Figure~\ref{fig:trainingsamples}. Please note that all these images are unprocessed. As shown in Figure~\ref{fig:dataset_comp}, we also conduct qualitative comparisons with other human datasets to demonstrate the superiority of our clean, high-quality data. Besides, we display more generated human images from the baseline model trained with our SHHQ in Figure~\ref{fig:concat_samples}.

\section{Experiment Results}
\label{sec:appendix_expresults}
\begin{figure}
  \centering
  \includegraphics[width=0.8\linewidth]{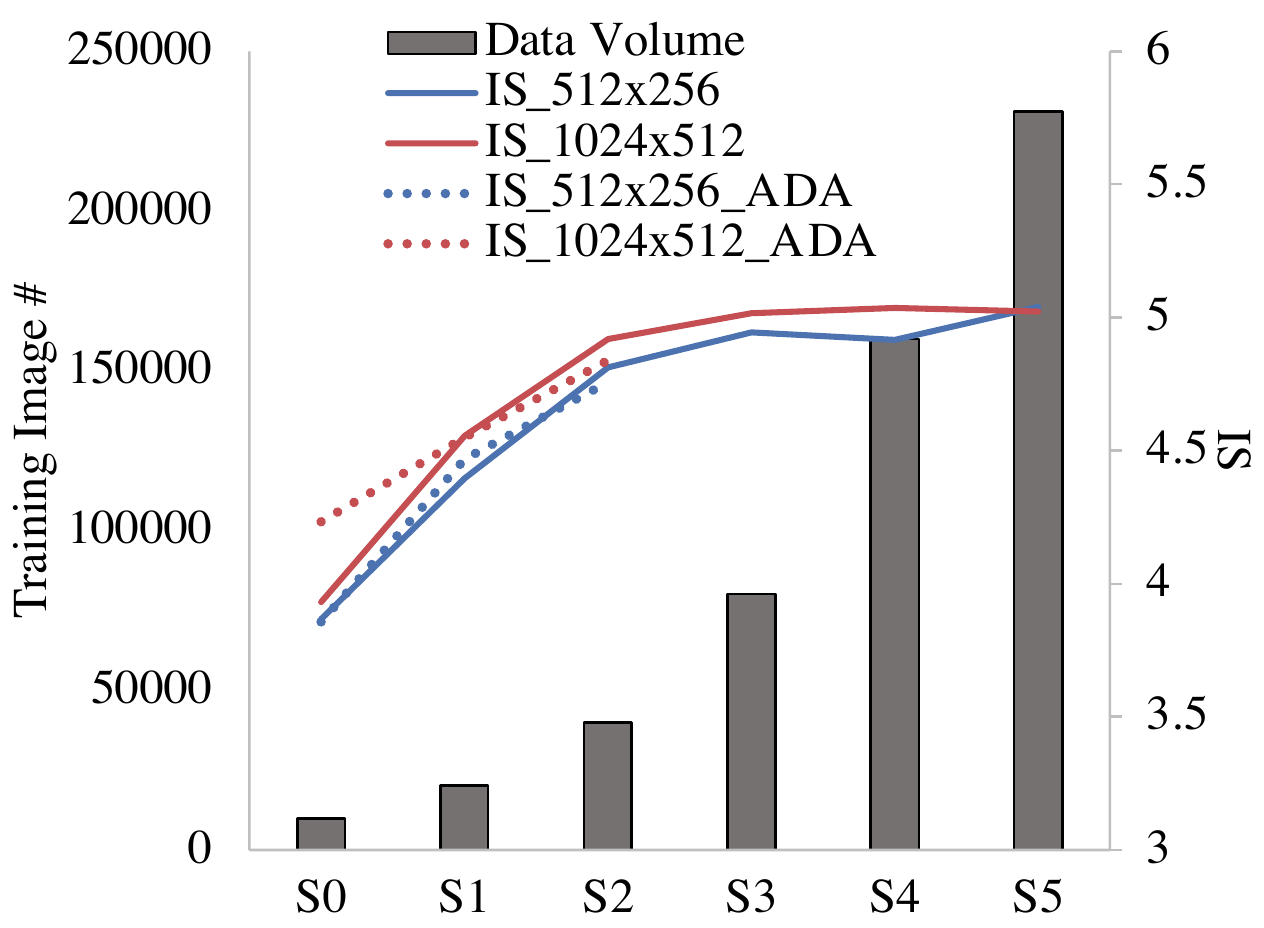}
  \caption{\textbf{IS scores.} IS scores for experiments $S0$ - $S5$ in $1024\times512$ and $512\times256$ resolutions. Dotted lines represents the IS scores with ADA strategies.} 
  \label{fig:isscore}
\end{figure}

\begin{figure*}
  \centering
  \includegraphics[width=0.95\textwidth]{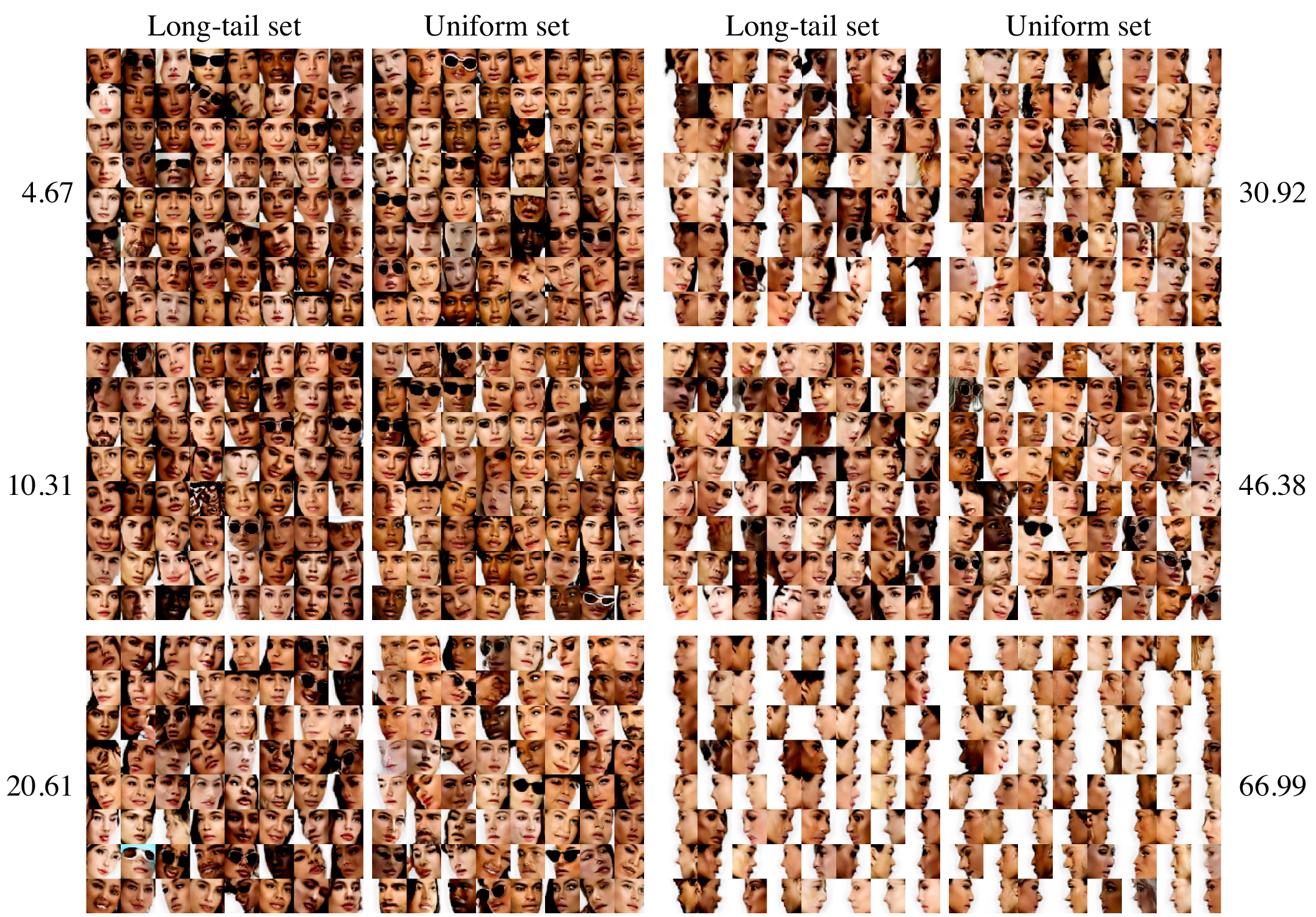}
  \caption{\textbf{Cropped faces with different face yaw angles from each bin.} All the images are generated from the long-tail and uniform experiments.}
  \label{fig:face_concat}
\end{figure*}

\begin{figure*}[h]
  \centering
  \includegraphics[height=0.9\textheight]{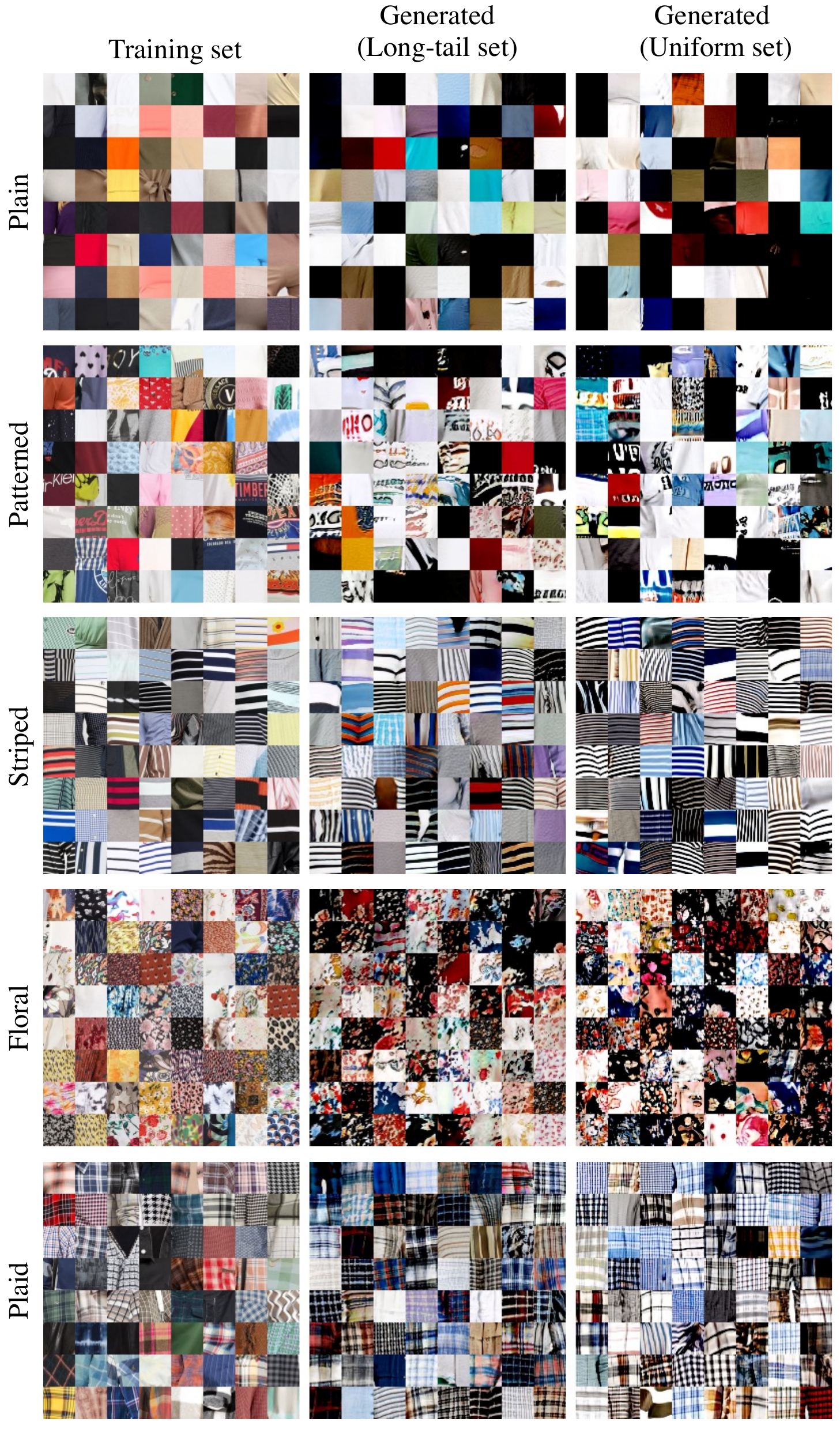}
  \caption{\textbf{Random cropped texture patches from each bin for both long-tail and uniform experiments.}}
  \label{fig:texture_concat}
\end{figure*}

Table~\ref{tab:fid_is} and Figure~\ref{fig:isscore} display the results of data size experiment elaborated in Section~\ref{sec:datasize}. The results align with our expectation that increment training data will improve IS scores and reduce FID scores. Figure~\ref{fig:face_concat} and~\ref{fig:texture_concat} depict the comparison between cropped faces and textures generated by the long-tail and uniform experiments. Due to privacy concerns, cropped training faces are not shown.

\begin{table}[h]
\begin{center}
  \caption{\textbf{FID and IS for experiments of data size.} Quantitative comparisons at resolutions of $512\times256$ and $1024\times512$ .}
  \label{tab:fid_is}
  \begin{tabular}{l|c|cc|cc}
    \toprule
         &           & \multicolumn{2}{c|}{$512\times256$} & \multicolumn{2}{c}{$1024\times512$}\\ 
         & Data Size &  FID &  IS  & FID  & IS  \\
    \midrule
    $S0$ & $10K$  & $7.80$ & $3.87$ & $7.23$ & $3.93$ \\    
    $S1$ & $20K$  & $4.46$ & $4.40$ & $4.33$ & $4.56$ \\
    $S2$ & $40K$  & $2.61$ & $4.81$ & $2.80$ & $4.92$ \\
    $S3$ & $80K$  & $2.53$ & $4.90$ & $2.09$ & $5.01$ \\  
    $S4$ & $160K$ & $2.09$ & $4.92$ & $2.02$ & $5.04$ \\  
    $S5$ & $230K$ & $1.97$ & $5.04$ & $1.57$ & $5.02$ \\      
    \bottomrule
  \end{tabular}
\end{center}
\end{table}

\begin{figure*}
  \centering
  \includegraphics[width=\linewidth]{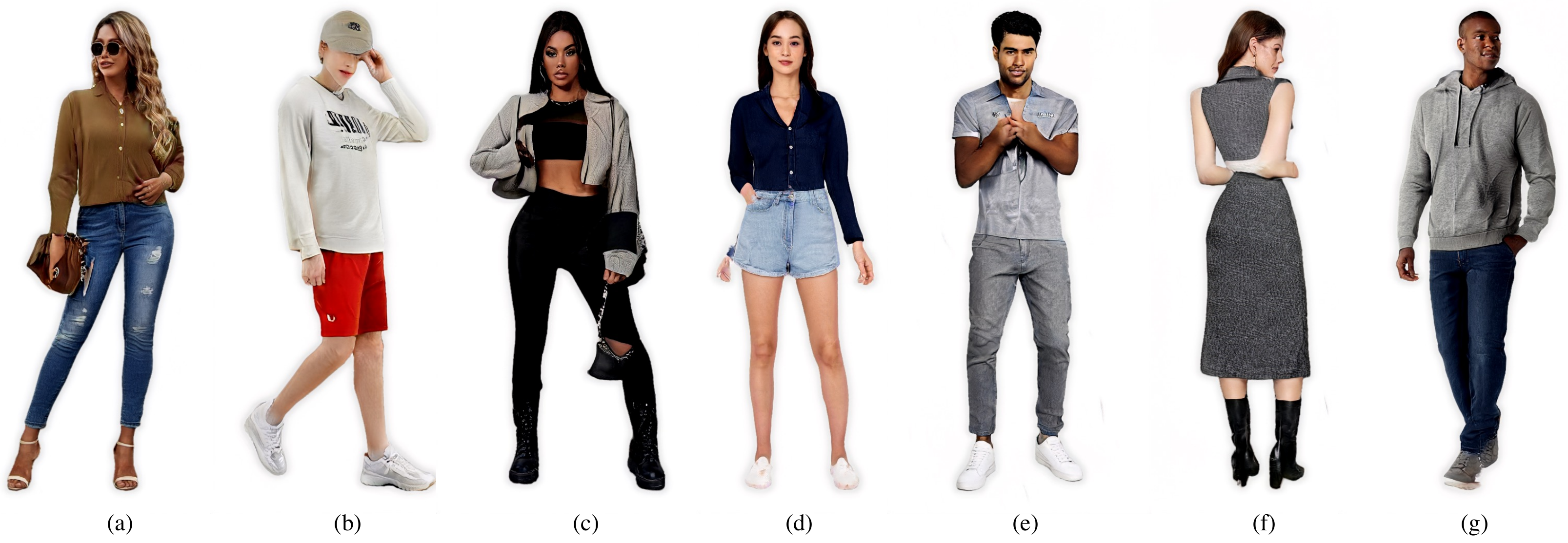}
  \caption{\textbf{Failure cases from baseline model.} (a) - (c): Features such as face, texture and accessories are entangled. (d) Three hands detected on a single person. (e) - (f): Inferior generated hand quality. (g): Face quality could be better.}
  \label{fig:failure_cases}
\end{figure*}

\section{Training Scheme}
\label{sec:appendix_training_scheme}
We adopt the official NVIDIA Pytorch version of StyleGAN2-ADA as our codebase, and use the architecture of StyleGAN2. Here are several settings we use to accommodate this human generation task: (a) The input human-image has a width-to-height ratio of $1:2$, and the input resolution in the script is changed accordingly. (b) We adopt the same eight mapping layers as the original StyleGAN~\cite{stylegan}. (c) There is no such a pretrained model for human images, so all the experiments are trained from scratch with the corresponding subset. (d) All other training hyper-parameters adopt the default values.

\section{Limitations}
\label{sec:appendix_limitations}
Compared to face generation, training an unconditional human GAN is an arduous task because the semantic features of the full-body are much more complicated than a single face. Figure~\ref{fig:failure_cases} shows some failure cases generated by the baseline model, suggesting several directions that can be strengthened in future human generation work. Artifacts caused by entangled features of faces/hands and clothing accessories are revealed in Figures~\ref{fig:failure_cases} (a) - (c). Case (d) exhibits three hands on a person, which indicates that the global perception of the model needs to be improved~\cite{huang2017beyond}. We observe inferior hand quality in rare poses such as (e) and (f). To address this, the potential work could be augmenting training with such extreme poses, changing the data distribution, or implementing independent networks (i.e. fine-grained discriminators) to enhance local details~\cite{detailmemore}. The face and texture quality in cases (b) and (g) could be enhanced by local refinement as well.

\section{Visualization of the Applications}
\label{sec:appendix_visualization}
\subsection{Style-Mixing}
\label{sec:appendix_stylemixing}
\begin{figure*}[hbtp]
  \centering
  \includegraphics[width=0.9\linewidth]{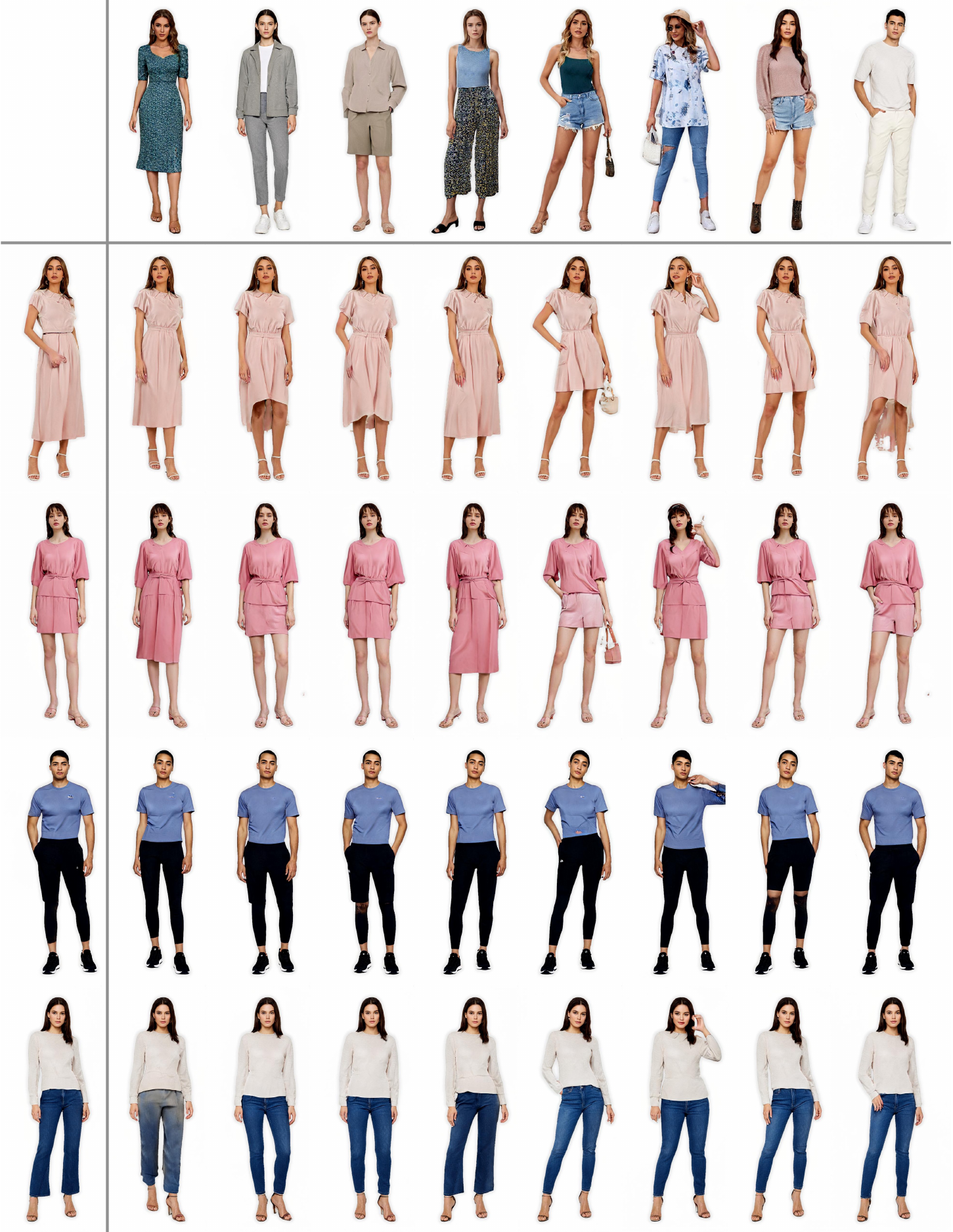}
  \caption{\textbf{Style-mixing} with copying styles of \textit{coarse} resolutions from reference images (top row), and rest spatial information are used from source images (first column).   }
  \label{fig:stylemixing_concat_low}
\end{figure*}

\begin{figure*}[hbtp]
  \centering
  \includegraphics[width=0.9\linewidth]{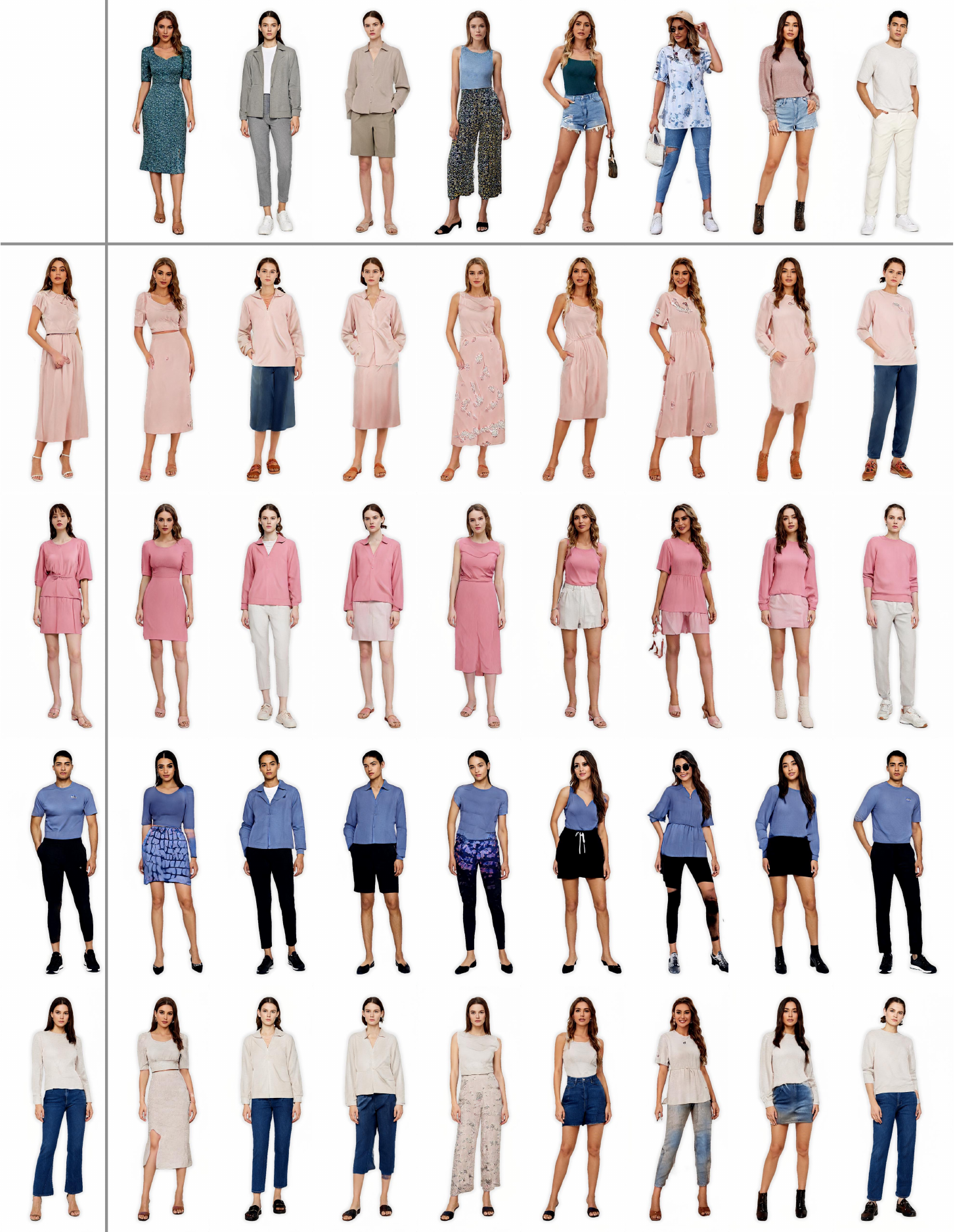}
  \caption{\textbf{Style-mixing} with copying styles of \textit{middle} resolutions from reference images (top row), and rest spatial information are used from source images (first column). }
  \label{fig:stylemixing_concat_mid}
\end{figure*}

\begin{figure*}[hbtp]
  \centering
  \includegraphics[width=0.9\linewidth]{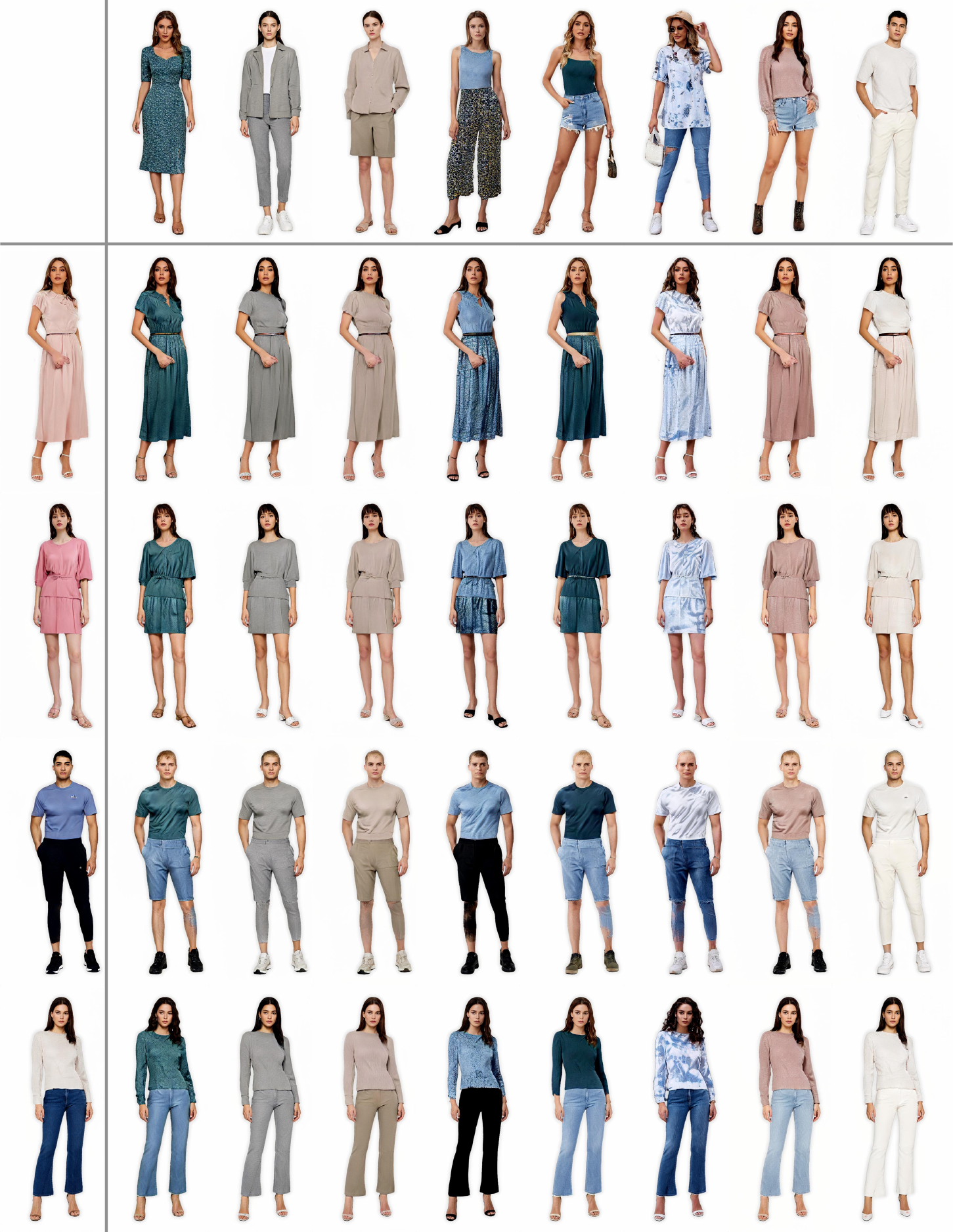}
  \caption{\textbf{Style-mixing} with copying styles of \textit{fine} resolutions from reference images (top row), and rest spatial information are used from source images (first column).}
  \label{fig:stylemixing_concat_high}
\end{figure*}

Here we provide more examples of images generated by style-mixing on our baseline model. Figure~\ref{fig:stylemixing_concat_low},~\ref{fig:stylemixing_concat_mid}, and~\ref{fig:stylemixing_concat_high} represent the results of style-mixing on coarse, middle, and high resolution respectively. It shows that the latent at different scales control different high-level attributes of the clothed human, which is similar to face images.


\subsection{Human Editing}
\label{sec:appendix_editing}
\begin{figure*}[htbp]
  \centering
  \includegraphics[width=0.9\linewidth]{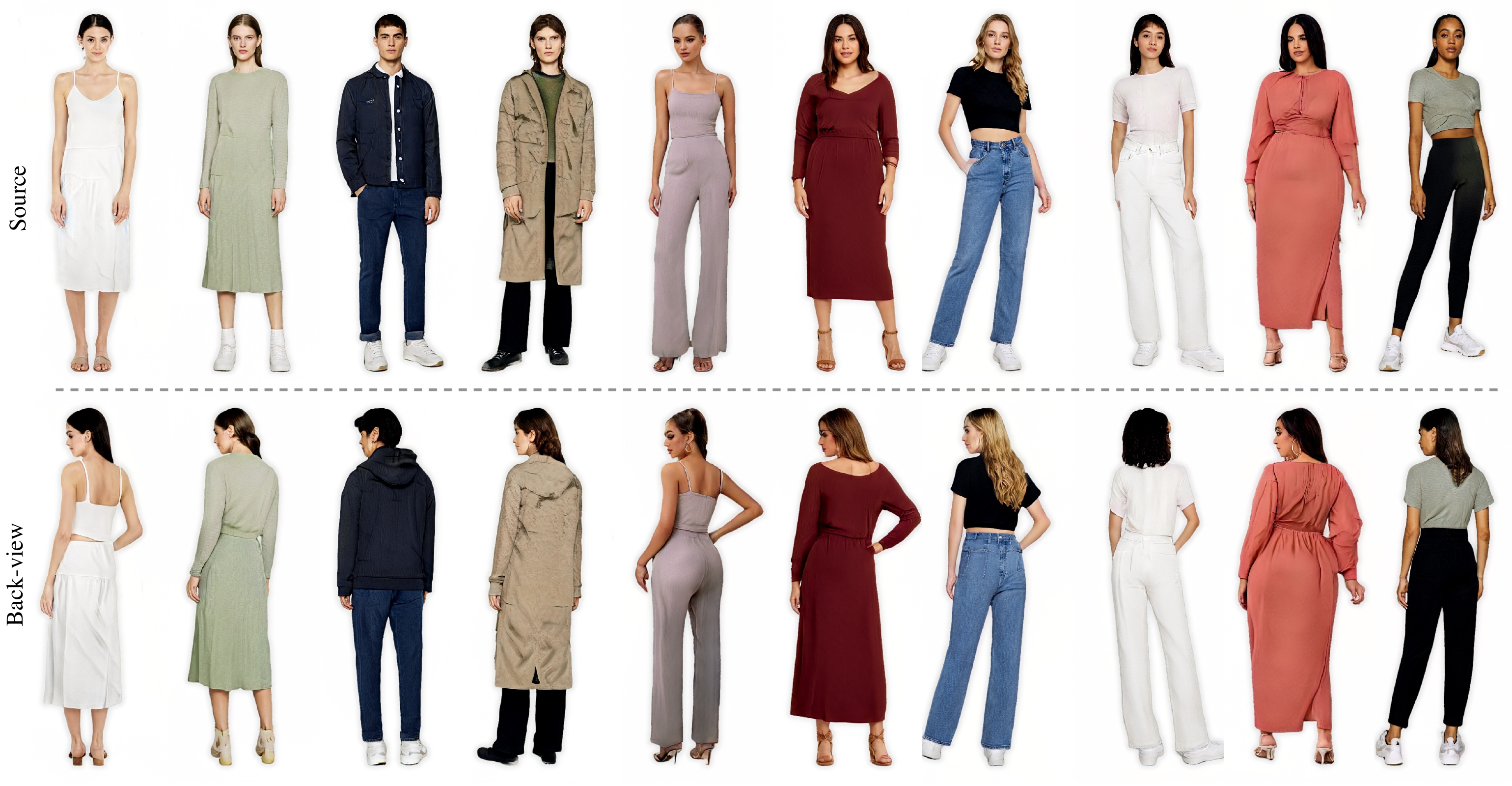}
  \caption{\textbf{Human Editing on orientation.}}
  \label{fig:editing_orientation}
\end{figure*}

\begin{figure*}[htbp]
    \centering
    \includegraphics[width=0.9\linewidth]{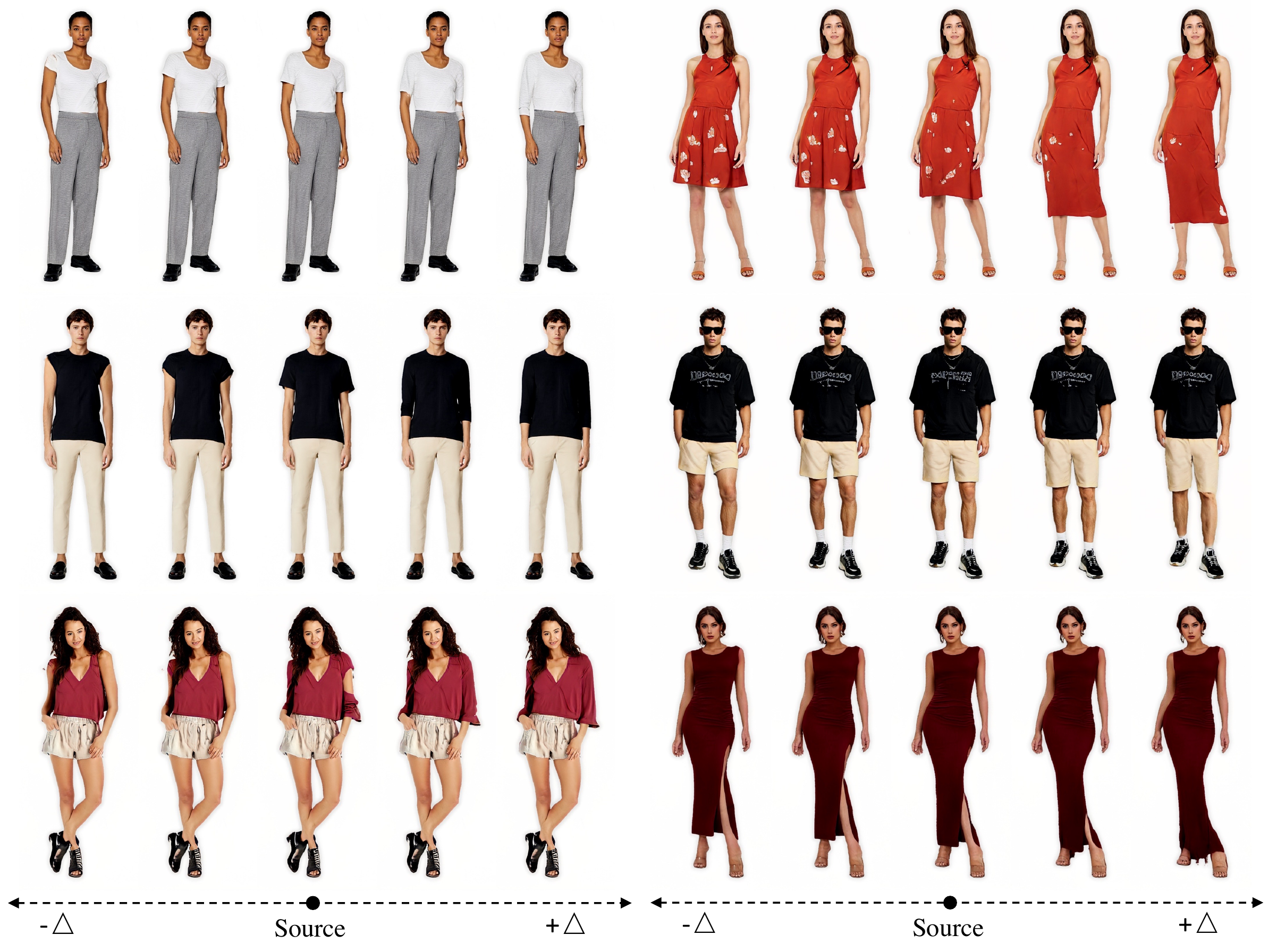}
    \caption{\textbf{Human Editing on human sleeve length (left) and bottom length (right).}}
    \label{fig:editing_sleeve_bottom}
\end{figure*}

Figure \ref{fig:editing_orientation} displays the rotation of the human from the front view to the back view. The editing is done in $W$ space.
Figure \ref{fig:editing_sleeve_bottom} demonstrates the editing results in the length of sleeves and bottoms, based on StyleSpace~\cite{wu2021stylespace}. 

